\documentclass[journal]{IEEEtran}
\usepackage{array}
\usepackage[caption=false,font=footnotesize]{subfig}
\usepackage{url}
\usepackage{graphicx}
\usepackage{cite}
\usepackage{booktabs}
\usepackage{multirow}
\usepackage{amsmath,amssymb,amsfonts}
\usepackage{color}
\usepackage{xcolor}
\usepackage{colortbl}
\usepackage{tabularx}

\usepackage{hyperref}
\newcommand{\reffig}[1]{Figure~\ref{#1}}
\newcommand{\refsec}[1]{Section~\ref{#1}}
\newcommand{\reftbl}[1]{Table~\ref{#1}}
\newcommand{\refequ}[1]{Equation~\ref{#1}}
\usepackage{tikz}
\usetikzlibrary{shapes.geometric, arrows, arrows.meta, positioning, calc, shadows, backgrounds, fit}
\usepackage[T1]{fontenc}

\usepackage{algorithm}
\usepackage{algpseudocode}
\begin{document}

\title{PedNStream: Scalable Network Flow Simulation for Pedestrian Traffic Management}

\author{Weiming Mai\IEEEauthorrefmark{1}, Dorine Duives and Serge Hoogendoorn
        % <-this % stops a space
\thanks{Weiming Mai, Dorine Duives and Serge Hoogendoorn are with the Department of Transportation and Planning, Delft University of Technology, Delft, Netherlands. (e-mail: w.m.mai@tudelft.nl; d.c.duives@tudelft.nl;s.p.hoogendoorn@tudelft.nl).
}
\thanks{\IEEEauthorrefmark{1}: Corresponding author.}
% \thanks{Manuscript received July 22, 2026.}
}

% \markboth{Manuscript submitted to IEEE transactions on intelligent transportation systems, 2026}%


\maketitle

\begin{abstract}
Evaluating operational crowd management at network scale requires simulations that can be run repeatedly while adapting interventions to changing conditions. Microscopic
models can represent detailed individual movement, but their computational cost
may limit their use in such repeated, network-scale evaluations. This paper
presents \emph{PedNStream} (\textbf{Ped}estrian \textbf{N}etwork Flow
\textbf{S}imulation), an open-source, Python-native simulator for macroscopic
pedestrian network simulation based on the Link Transmission Model (LTM).
\emph{PedNStream} extends LTM-based pedestrian models with stochastic link
dynamics that represent local variation in pedestrian flow. It uses a
utility-based route-choice model to capture how pedestrians adjust their route
choices in response to congestion and control interventions as conditions
change over time. The modular framework provides controller interfaces for gating, flow
separation, and route guidance. We evaluate \emph{PedNStream} in a staged manner. Synthetic scenarios verify
key crowd-dynamics mechanisms, including queue formation, spillback, congestion
dissipation, and adaptive rerouting. Real-network experiments assess
large-scale behavior against observed pedestrian counts. A closed-loop case
study demonstrates controller integration, and a runtime analysis quantifies
scalability. These results position \emph{PedNStream} as an efficient and practical testbed for large-scale pedestrian network simulation and crowd management research.
\end{abstract}


\section{INTRODUCTION}
\IEEEPARstart{E}{ffective} crowd management during large events ensures pedestrian safety and prevents congestion. Simulation methods have been developed over the years to model crowd dynamics to aid infrastructure design and tactical management (e.g., fencing and capacity limits). Broadly, these methods fall into three categories: microscopic, mesoscopic \cite{helbing1995social,blue1998emergent,daamen2004modelling, hoogendoorn2004pedestrian, Rasouli2021PedestrianSA}, and macroscopic formulations that either model pedestrian flow on networks \cite{lovaas1994modeling, chalmet1982network,van2020estimation} or describe pedestrian movement as spatial-temporal continuous flow \cite{hoogendoorn2015continuum, duives2016continuum}. The third category is mesoscopic model, it provides a compromise between the level of detail of the flow description and computational complexity for the simulation of large scale scenarios \cite{tordeux2018mesoscopic}. Most existing simulation models for pedestrian dynamics serve a descriptive purpose: they assist practitioners and policymakers in evaluating the effects of infrastructural changes. These tools provide insights into how crowds behave under specific conditions, but they are typically designed for offline scenario evaluation rather than real-time, operational-level control, where network states must be repeatedly observed and interventions updated during the simulation.

At the operational level, Hänseler et al. proposed an optimal crowd-control approach for a metro station \cite{hanseler2018optimal}, using dynamic reductions of passenger flows at check-in gates and dynamic escalator reversal. More recently, Molyneaux et al. \cite{molyneaux2022controlling, molyneaux2021design} introduced the concept of a Dynamic Pedestrian Management System (DPMS), focusing on localized interventions such as flow separators, moving walkways, and gates in corridors, intersections, and entrances/exits. These studies demonstrate the value of control-oriented pedestrian modeling, but they mainly address local or facility-level interventions. For large-scale events, evacuations, festivals, or urban crowd-management settings, there remains a need for simulation tools that can evaluate control strategies repeatedly over large pedestrian networks in a rolling-horizon framework.

Meeting this need requires a model that combines computational efficiency with a realistic representation of pedestrian movement. High-resolution approaches, such as agent-based microscopic models and spatially discretized continuum models \cite{duives2016continuum,hanseler2017dynamic}, can capture detailed movement patterns but are often too computationally expensive for repeated simulations over large networks. Aggregate network models offer greater efficiency, including discrete utility-based models \cite{borgers1986model}, network flow optimization approaches \cite{chalmet1982network}, and queueing-theoretic representations \cite{lovaas1994modeling}. However, these models were developed primarily for planning, evacuation analysis, or route-demand assignment rather than the repeated evaluation of operational control strategies. The central challenge is therefore to preserve the scalability of network models while extending them to support realistic and responsive pedestrian crowd-management applications.

Recently, Lilasathapornkit \& Saberi.~\cite{lilasathapornkit2022dynamic} developed an LTM-based dynamic pedestrian traffic assignment model for bidirectional sidewalk networks. Although the model supports large-scale simulation, its route-choice mechanism is based on dynamic user equilibrium (DUE), which assumes that pedestrians adapt toward consistent route choices based on experienced travel conditions. This assumption is more plausible in familiar or recurrent settings than in environments characterized by unfamiliar visitors, rapidly changing conditions, or active crowd-management interventions. In such cases, route choices may be more heterogeneous, stochastic, and directly influenced by guidance or control actions.

\begin{table*}[!htp]
    \caption{Positioning of {\normalfont\bfseries PedNStream} relative to representative open-source simulation tools. The comparison highlights the trade-off between modeling resolution and application scale. {\normalfont\itshape Limited}: Control is not a primary use case; {\normalfont\itshape Indirect}: Control is feasible via external APIs/wrappers; {\normalfont\itshape Direct}: Native formulation for iterative, learning-based, or real-time control optimization.}
    \label{tab:pednstream_positioning}
    \centering
    \footnotesize
    \renewcommand{\arraystretch}{1.4}
    \setlength{\tabcolsep}{6pt}
    
    \begin{tabularx}{\textwidth}{l l l X l}
        \toprule
        \textbf{Tool} & \textbf{Level} & \textbf{Typical Scale} & \textbf{Primary Modeling Focus} & \textbf{Closed-loop Control} \\
        \midrule
        JuPedSim~\cite{wagoum2015jupedsim} & Microscopic & Facility / Local & Individual locomotion & Limited \\
        \addlinespace[2pt]
        Vadere~\cite{kleinmeier2019vadere} & Microscopic & Facility / Local & High-fidelity crowd/evacuation dynamics & Limited \\
        \addlinespace[2pt]
        SUMO~\cite{lopez2018microscopic} & Micro/Meso & Urban / Network & Multimodal traffic & Indirect \\
        \addlinespace[2pt]
        MATSim~\cite{w2016multi} & Micro/Meso & Urban / Regional & Large-scale travel demand & Indirect \\
        
        \midrule
        \rowcolor[gray]{0.95}
        \textbf{PedNStream} & \textbf{Macroscopic} & \textbf{Urban / Network} & \textbf{Flow-based network dynamics for control optimization} & \textbf{Direct} \\
        \bottomrule
    \end{tabularx}
\end{table*}
The model also retains several assumptions inherited from vehicular LTM formulations. It does not explicitly represent stochastic pedestrian behaviors, such as temporary stops for activities, nor variations in walking speed across individuals or groups. These omissions may lead to overly regular flow patterns and limit the model's ability to reproduce the variability observed in operational crowd environments.

To address these limitations, we developed \emph{PedNStream}, an open-source Python framework for efficient, network-scale pedestrian simulation and control-oriented experimentation. Designed for urban networks and large event settings, the framework supports the repeated evaluation of crowd-management interventions, including crowd redirection, capacity adjustment, and flow separation. As summarized in \reftbl{tab:pednstream_positioning}, \emph{PedNStream} complements existing high-resolution pedestrian simulators and large-scale travel-demand models by focusing specifically on macroscopic pedestrian-network dynamics and closed-loop crowd management.

% The contributions of this paper are threefold. First, we extend the LTM-based pedestrian model of Lilasathapornkit et al.~\cite{lilasathapornkit2022dynamic} by adding stochastic and diffusive link dynamics~\cite{liu2015modeling} and by replacing DUE-based route choice with a utility-based formulation for heterogeneous and adaptive route responses. Second, we implement these modeling components in \emph{PedNStream}, a modular, open-source, and control-oriented simulator for macroscopic pedestrian networks. The framework includes controller interfaces and two baseline gate controllers, and its source code and installable Python package are publicly available~\cite{Weiming_Mai_PedNStream_2026}. Third, we evaluate the framework through staged experiments covering mechanism verification, link-dynamics comparison, real-network case studies, closed-loop controller integration, and runtime scalability.

This paper makes three contributions. First, it extends the LTM-based pedestrian model of Lilasathapornkit et al.~\cite{lilasathapornkit2022dynamic} with stochastic, diffusive link dynamics~\cite{liu2015modeling} and a utility-based route-choice formulation for heterogeneous, adaptive responses. Second, it implements these components in \emph{PedNStream}, an open-source, modular simulator for macroscopic pedestrian networks with controller interfaces and baseline gate controllers.\footnote{The source code, documentation, and installation instructions are publicly available at~\cite{Weiming_Mai_PedNStream_2026}.} Third, it evaluates the framework through synthetic mechanism tests, partial real-network count reconstruction, closed-loop controller experiments, and runtime analysis.

The remainder of this paper is organized as follows. \refsec{dnl} presents the pedestrian link dynamics. \refsec{route_node} describes route choice and node flow assignment. \refsec{architecture} introduces the simulation framework and controller interface. \refsec{syn_exp} reports the synthetic scenario evaluation, and \refsec{real_network} presents the real-network cases. \refsec{control_runtime} contains the crowd-management case study and runtime analysis. Finally, \refsec{conclusion} concludes the paper. The appendices describe the software architecture and the baseline gate controllers.

% \section{Dynamic Network Loading Problem}\label{dnl}
% Dynamic Network Loading (DNL) simulates how traffic propagates through a network over time. It is a core component of Dynamic Traffic Assignment (DTA) frameworks, which model route choice under time-varying congestion. In this section, we introduce the numerical solution of the DNL problem, with particular emphasis on the Link Transmission Model (LTM). Table~\ref{tab:notation} summarizes the main notation used throughout the model description.

\section{LTM-Based Pedestrian Link Dynamics}\label{dnl}

This section presents the link-level pedestrian-flow dynamics used in \emph{PedNStream}. We first introduce the standard Link Transmission Model (LTM), a link-based cumulative-flow representation of the first-order LWR kinematic-wave model and flow-conservation principle. We then review the extension of LTM to bidirectional pedestrian movement proposed by Lilasathapornkit \& Saberi~\cite{lilasathapornkit2022dynamic}. Building on these formulations, we introduce the link-model refinements implemented in \emph{PedNStream}, including time-varying travel times, width and area-based capacity constraints, occupancy-dependent boundary demand, and stochastic pedestrian flow dynamics. Together, these components determine the sending and receiving flows of each link, while route choice and node level flow assignment are described in the subsequent sections. Table~\ref{tab:notation} summarizes the main notation used throughout the model description.

\begin{table}[!t]
\centering
\caption{Main notation used in the model.}
\label{tab:notation}
\renewcommand{\arraystretch}{1.05}
\setlength{\tabcolsep}{4pt}
\begin{tabularx}{\columnwidth}{>{\raggedright\arraybackslash}p{0.38\columnwidth} >{\raggedright\arraybackslash}X}
\toprule
\textbf{Symbol} & \textbf{Definition} \\
\midrule
$t,\,\Delta t$ & Time index and step size. \\
$i, j$ & Link indices (opposite directions when paired). \\
$n$ & Node index. \\
$U_i(t),\,V_i(t)$ & Cumulative inflow and outflow on link $i$. \\
$S_i(t),\,R_i(t)$ & Sending and receiving flows on link $i$. \\
$C_i,\,L_i,\,w_i,A_i$ & Capacity, length, width and area of link $i$. \\
$T_f,\,T_{\omega},\,T(t)$ & Free-flow, shockwave, and realized travel time in physical units. \\
$\tau_f,\,\tau_{\omega},\,\tau(t)$ & Corresponding travel-time delays in simulation steps. \\
$k_i,\,k_j$ & Directional densities. \\
$k_{\mathrm{jam}},\,k_c$ & Jam and critical density. \\
% $\rho_i$ & Density ratio, $\rho_i = \frac{k_i}{k_i+k_j}$. \\
$X_i(t)$ & Number of pedestrians withheld on link $i$ by activity at time $t$. \\
$\mathcal{U}_{\ell}$ & Utility of downstream link $\ell$. \\
$P(\ell_j \mid \ell_i),\, p_{ij}$ & Turning fraction from link $i$ to $j$. \\
$q_{ij}^n$ & Flow from $i$ to $j$ at node $n$. \\
$F,\,\gamma,\,p_{\mathrm{rel}},\,p_{\mathrm{activity}}$ & Diffusion coefficient, diffusion calibration parameter, release probability, and activity probability. \\
% $S_i^n,\,R_j^n$ & Node-level sending and receiving capacities. \\
\bottomrule
\end{tabularx}
\end{table}

\subsection{The Link Transmission Model}\label{ltm}
Yperman et al.~\cite{yperman2005link} proposed the LTM model, a dynamic network-loading model for vehicular traffic, derived from first-order kinematic-wave theory~\cite{lighthill1955kinematic} and the principle of flow conservation. LTM consists of two coupled components: a link model that captures the evolution of traffic states within each link, and a node model that determines the admissible transfer of flow between connected links.

At each time step, the link model determines how much flow can leave a link (sending flow, i.e., upstream demand) and how much flow can enter a link (receiving flow, i.e., downstream supply). The actual transfer across a link boundary is therefore limited by both quantities. Let $T_f$ and $T_{\omega}$ denote the free-flow and shockwave travel times in physical units, respectively. Their corresponding integer delays are $\tau_f=\mathrm{round}(T_f/\Delta t)$ and $\tau_{\omega}=\mathrm{round}(T_{\omega}/\Delta t)$. For a simulation step of length $\Delta t$, the sending flow from an upstream link $i$ over the time interval $(t-\Delta t, t]$ is computed as follows:
\begin{subequations}
\begin{align}
    &S_{i,\text{boundary}}(t) = U_i(t - \tau_{f}) - V_i(t - \Delta t), \label{eq:S_boundary} \\
    &S_{i,\text{link}}(t) = C_i \Delta t, \label{eq:S_link} \\
    &S_i(t) = \min(S_{i,\text{boundary}}(t), S_{i,\text{link}}(t)). \label{eq:S_total}
\end{align}
\end{subequations}
Here, $U_i(t)$ and $V_i(t)$ denote the cumulative inflow and outflow of link $i$ up to time $t$, respectively, and $C_i$ is the flow capacity of the link. In the standard LTM, the capacity is treated as a fixed link-level bound. The term $S_{i,\text{boundary}}(t)$ represents the amount of flow that reaches the downstream boundary of link $i$ after the free-flow delay $\tau_f$, whereas $S_{i,\text{link}}(t)$ imposes the link-capacity limit. Hence, the sending flow $S_i(t)$ is the minimum of these two quantities. Correspondingly, the receiving flow is written as
\begin{subequations}
\begin{align}
    &R_{i,\text{boundary}}(t) = V_i(t - \tau_{\omega}) +k_{\mathrm{jam}}L_i - U_i(t - \Delta t), \label{eq:R_boundary} \\
    &R_{i,\text{link}}(t) = C_i \Delta t, \label{eq:R_link} \\
    &R_i(t) = \min(R_{i,\text{boundary}}(t), R_{i,\text{link}}(t)), \label{eq:R_total}
\end{align}
\end{subequations}
where $R_{i,\text{boundary}}(t)$ represents the space available for new inflow at the upstream boundary of link $i$. It is determined by three terms: the cumulative outflow shifted by the shockwave delay $\tau_{\omega}$, the maximum storage $k_{\mathrm{jam}}L_i$, and the cumulative inflow already admitted into the link. The term $R_{i,\text{link}}(t)$ imposes the link-capacity limit. Here, $k_{\mathrm{jam}}$ denotes the jam density, i.e., the density at which the link is fully occupied and pedestrian movement becomes negligible.

\subsection{Link Transmission Model for Pedestrian Dynamics}
The standard LTM provides the cumulative-flow and demand--supply framework used to propagate flow through the network. However, its conventional vehicular formulation assumes fixed free-flow travel times and does not explicitly represent interactions between opposing pedestrian streams. Therefore, Lilasathapornkit et al.~\cite{lilasathapornkit2022dynamic} proposed a bidirectional pedestrian LTM that adapts the standard formulation to account for these interactions. 

% Lilasathapornkit et al. \cite{lilasathapornkit2022dynamic} extend the standard Link Transmission Model (LTM) to incorporate bi-directional interactions. 
% In their formulation, 
Instead of using the constant free-flow speed, they introduce the effective free-flow speed $\hat{v}_f(t)$ of a link $i$ that is influenced by the density of its counter-directional link $j$. This change allows the model to adapt to actual traffic conditions, and travel time is no longer a constant. Another adjustment modifies the boundary conditions, particularly the receiving flow, as shown in \refequ{lilaetal}. The key idea is that the number of pedestrians entering a downstream link is constrained not only by congestion within the link but also by the opposing flow from the counter-directional link. The modified components are highlighted in blue in the following equations:
\begin{subequations}
\begin{align}
S_{i,\text{boundary}}(t) 
&= U_i(t - \textcolor{blue}{\hat{\tau}_f(t)}) - V_i(t - \Delta t), \label{talink} \\
R_{i,\text{boundary}}(t) 
&= V_i(t - \tau_{\omega}) + k_{\mathrm{jam}}L_i \notag \\
&\quad - U_i(t - \Delta t) - \textcolor{blue}{S_{j}(t)} .
\label{lilaetal}
\end{align}
\end{subequations}
In the above equations, $S_j(t)$ denotes the sending flow of the counter-directional stream $j$ at time step $t$. The effective free-flow travel time is written as $\hat{T}_f(t)=L_i/\hat{v}_f(t)$ in physical time units and $\hat{\tau}_f(t)=\mathrm{round}(\hat{T}_f(t)/\Delta t)$ in simulation steps. To calculate the effective free-flow speed $\hat{v}_f(t)$, they first calculate $\rho_i$, as the ratio of the density of the reference direction $i$ to the density of the opposite direction $j$ in a bidirectional stream: $\rho_i(t) = \frac{k_i(t)}{k_i(t) + k_j(t)}$. This term captures the relative dominance of the forward-direction flow. 

The effective free-flow speed is then calculated as $\hat{v}_f(t) = \rho_i^{\lambda}(t) v_f$, where $v_f$ is the nominal free-flow speed of the link $i$, $k_i$ and $k_j$ are the densities of both directions, respectively. $\lambda$ is a calibration parameter. Although this formulation produces a time-varying free-flow speed, it depends on the relative directional intensities rather than the total link density. Consequently, identical directional proportions yield the same effective free-flow speed at different congestion levels, while unidirectional flow retains the nominal free-flow speed regardless of its density.

\subsection{Refinements to the Pedestrian LTM}
To address these limitations and better represent pedestrian movement, \emph{PedNStream} introduces four refinements to the bidirectional pedestrian LTM: (i) a boundary sending flow based on realized travel time, which responds to total within-link congestion rather than only the relative directional intensities; (ii) a width-dependent discharge capacity that represents the effect of usable walking width; (iii) an area and occupancy-based receiving constraint that accounts for two-dimensional shared space; and (iv) a stochastic sending constraint that represents variability in walking speeds, local interactions, and pedestrian activities. In the equations below, purple denotes components introduced in \emph{PedNStream}, whereas blue denotes the opposing-flow term retained from the model of Lilasathapornkit et al.

\paragraph{Time-varying boundary sending flow.}
We first replace the effective free-flow travel time in \refequ{talink} with the time-varying realized travel time $T(t)$. Expressed in simulation steps, this delay is $\tau(t)=\mathrm{round}(T(t)/\Delta t)$. The corresponding raw boundary sending flow is
\begin{equation}
\tilde{S}_{i,\mathrm{boundary}}(t)
=U_i(t-\textcolor{purple}{\tau(t)})-V_i(t-\Delta t),
\label{eq:time_varying_boundary}
\end{equation}
which allows the sending boundary to respond to the prevailing within-link congestion state rather than relying on a directional-ratio-based effective free-flow travel time.

This time-varying delay creates a practical issue during the initial simulation stage and under severe congestion. In these cases, $\tau(t)$ may exceed the elapsed horizon, making the delayed index $t-\tau(t)$ non-positive and the raw cumulative-count difference in \refequ{eq:time_varying_boundary} unreliable. We therefore compute $T(t)$ as a moving average of the instantaneous link travel time, so that $\tau(t)$ changes gradually, and clip $\tilde{S}_{i,\mathrm{boundary}}(t)$ to remain non-negative. We then blend this delayed demand with the current link occupancy using
\begin{equation}
\xi_i(t)=\mathrm{clip}\!\left(
\frac{k(t)-k_c}{k_{\mathrm{jam}}-k_c},0,1\right),
\label{eq:congestion_weight}
\end{equation}
where $k(t)$ is the current link density and $k_c$ is the critical density. The resulting boundary sending flow is
\begin{equation}\label{eq:bound}
S_{i,\mathrm{boundary}}(t)
=\xi_i(t)N_i(t)
+\bigl(1-\xi_i(t)\bigr)\tilde{S}_{i,\mathrm{boundary}}(t),
\end{equation}
where $N_i(t)$ is the number of pedestrians currently on link $i$. When $k(t)\leq k_c$, the model follows the delayed cumulative-count demand. As the density approaches $k_{\mathrm{jam}}$, it shifts smoothly toward the occupancy-based demand, thereby avoiding unrealistically low or unstable sending demand under severe congestion.

\paragraph{Width-dependent discharge capacity}
To represent the two-dimensional nature of pedestrian discharge, we interpret $C_i$ as the capacity of link $i$ per unit width. The link sending capacity is therefore
\begin{equation}
S_{i,\mathrm{link}}(t)
=C_i\,\textcolor{purple}{w_i}\,\Delta t,
\label{width}
\end{equation}
where $w_i$ is the usable discharge width. Thus, narrowing the available width reduces the number of pedestrians that can leave the link during a simulation step, whereas widening it increases the admissible discharge.

\paragraph{Area and occupancy-based receiving constraint}
Pedestrian movement occupies two-dimensional space, so downstream storage is determined by usable area rather than link length alone. We define the boundary receiving flow as
\begin{align}
R_{i,\mathrm{boundary}}(t)
&=V_i(t-\tau_{\omega})
+k_{\mathrm{jam}}\,\textcolor{purple}{A_i} \notag\\
&\quad-U_i(t-\Delta t)
-\textcolor{purple}{N_j(t)}
-\textcolor{blue}{S_j(t)},
\label{eq:ped_recv}
\end{align}
where $A_i$ is the usable area of link $i$, and hence $k_{\mathrm{jam}}A_i$ is its maximum pedestrian storage. The term $N_j(t)$ is the occupancy of the counter-directional link $j$ and accounts for the physical space already occupied by pedestrians moving in the opposite direction. The term $S_j(t)$ retains the opposing-flow boundary constraint from the preceding bidirectional formulation. Together, these terms prevent the admitted flow from exceeding the shared usable space.

\paragraph{Stochastic sending constraint}
Finally, we introduce $S_{i,\mathrm{stochastic}}(t)$ to represent variability arising from heterogeneous walking speeds, local interactions, and pedestrian activities. The realized sending flow is bounded by the boundary demand, the width-dependent capacity, and this stochastic constraint:
\begin{align}
S_i(t)
&=\min\Bigl(
S_{i,\mathrm{boundary}}(t),
S_{i,\mathrm{link}}(t), \notag\\
&\qquad\textcolor{purple}{S_{i,\mathrm{stochastic}}(t)}
\Bigr).
\label{eq:stochastic}
\end{align}
All sending and receiving quantities are clipped to remain non-negative. The next subsection describes how $S_{i,\mathrm{stochastic}}(t)$ is computed under free-flow and congested conditions.

\subsection{Stochastic Flow Modeling}
\begin{figure}[htb!]
    \centering
    \includegraphics[width=.9\columnwidth]{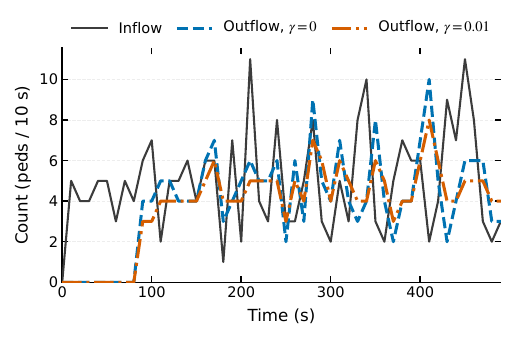}
    \caption{Illustration of the effect of the diffusion parameter $\gamma$ on downstream outflow under free-flow conditions. Larger $\gamma$ values produce a broader and lower-peaked outflow profile.}
    \label{fig:gamma}
\end{figure}

This subsection details the stochastic sending constraint introduced in \refequ{eq:stochastic}. The stochastic term is designed to capture variability in pedestrian movement that is not represented by the deterministic LTM alone. Under free-flow conditions (i.e. density lower than $k_c$), we use the crowd-diffusion model~\cite{liu2015modeling,mai2025learning} to represent dispersion in walking speeds. Let $t$ denote the current simulation-step index, let $\hat{q}_{\uparrow}(t)$ denote the upstream inflow available to traverse the link, let $\hat{q}_{\downarrow}(t)$ denote the resulting downstream outflow, and let $\tau(t)=\mathrm{round}(T(t)/\Delta t)$ be the realized link travel time expressed in simulation steps. The diffused downstream flow is then computed as
\begin{equation}
\hat{q}_{\downarrow}(t)
= \sum_{m=0}^{t-\tau(t)-1} F(t)\bigl(1-F(t)\bigr)^m
\, \hat{q}_{\uparrow}(t-\tau(t)-m).
\label{diffusion}
\end{equation}
where $F(t)=\frac{1}{1+\gamma T(t)}$ is the time-dependent diffusion coefficient, $\gamma$ is a calibration parameter, and $T(t)$ is the realized link travel time in physical units. This formulation distributes the release of a cohort over several subsequent time steps. Larger travel times or larger values of $\gamma$ therefore produce a more dispersed downstream outflow profile. As illustrated in \reffig{fig:gamma}, when $\gamma=0$, the downstream outflow is nearly a time-shifted copy of the inflow, whereas larger values of $\gamma$ flatten the peak and spread the discharge over a longer interval.

For $t\leq\tau(t)$, no cohort has completed the realized link traversal. The diffused downstream flow, and hence the stochastic sending flow in the free-flow regime, is set to zero.

Under congested conditions, \emph{PedNStream} shifts from the diffusion model to a stochastic release model. The free-flow diffusion model disperses cohort arrival times but does not represent the possibility that pedestrians who are eligible to leave a link may fail to discharge because of local crowd interactions, speed adjustments, or competition for limited exit space. We therefore treat each of the $n_i(t)=\lfloor S_{i,\mathrm{boundary}}(t)\rfloor$ eligible pedestrians as an independent Bernoulli release trial with probability $p_{\mathrm{rel}}$. The number released during step $t$ then follows a binomial distribution. Rather than treating this probability as fixed, we relate it to the congestion weight in \refequ{eq:congestion_weight}:
\begin{equation}
p_{\mathrm{rel}}(t)
=p_{\max}-\bigl(p_{\max}-p_{\min}\bigr)\xi_i(t),
\label{eq:release_probability}
\end{equation}
where $p_{\min}$ and $p_{\max}$ are user-configurable scenario bounds. Thus, the release probability equals $p_{\max}$ at the onset of congestion and decreases smoothly toward $p_{\min}$ as the density approaches $k_{\mathrm{jam}}$. In the experiments reported here, we use $p_{\min}=0.8$ and $p_{\max}=1$. This formulation bounds the realized outflow between $0$ and $n_i(t)$, captures both reduced discharge efficiency and step-to-step variability, and gives $p_{\mathrm{rel}}$ a direct interpretation as the conditional probability that an eligible pedestrian is released during the current time step.

In either regime, if activity behavior is enabled, a second binomial draw with probability $p_{\mathrm{activity}}$ is applied to the preliminary stochastic sending flow. This draw represents pedestrians who would otherwise be able to leave but remain on the link to perform activities. Sampling from the preliminary sending flow ensures that activity-related retention does not exceed the number otherwise available for discharge.

These parameters are not treated as universal constants. The diffusion coefficient $\gamma$ controls free-flow dispersion and can be calibrated to observed flow dynamics~\cite{liu2015modeling}. In contrast, the bounds $p_{\min}$ and $p_{\max}$ and the activity probability $p_{\mathrm{activity}}$ are scenario-level parameters that allow users to specify the density-dependent congested-flow discharge and activity-related retention. Users can vary these parameters to construct behavioral scenarios and evaluate the sensitivity of simulation or control outcomes. Algorithm~\ref{alg:sending_flow} summarizes the stochastic sending-flow calculation at a link.

% The smoothing factor $\delta$ plays a different role: it is a numerical stabilization parameter rather than a behavioral one, and in our implementation it is restricted to $\delta \in [0.5,1]$ so that the current-step estimate retains dominant weight while step-to-step oscillations are still damped.

\begin{algorithm}[H]
\caption{Stochastic sending flow update at link $i$}
\label{alg:sending_flow}
\begin{algorithmic}[1]
\Require Parameters $\Delta t$, $\gamma$, $k_c, k_{\text{jam}}$. Probability $p_{\min},p_{\max}\in[0,1]$, and $p_{\mathrm{activity}}\in[0,1]$
\Procedure{CalSendingFlow}{$t$}  \Comment{$t$ = current time step}
    % \State $\rho \gets \texttt{get\_total\_link\_density}(t)$
    \State $k(t) \gets k_i(t) + k_j(t)$
    % \State $\textcolor{purple}{\tau} \gets \text{round}\left( \frac{\text{avg\_travel\_time}[t]}{\Delta t} \right)$
    \State $\textcolor{purple}{\tau(t)} \gets \mathrm{round}\!\left(\dfrac{T(t)}{\Delta t}\right)$

    \If{$t < \tau_f$}
    \Comment{initial stage}
        % \State $S_{i}(t) \gets 0$
        \State \Return $S_{i}(t)\gets 0$
    \EndIf

    % \State calculate link congestion factor $c \gets \text{clip}\left(\frac{\rho_i}{k_{\text{jam}}}, 0, 1\right)$
    \State Compute $S_{i,\text{boundary}}(t)$ according to \refequ{eq:bound}.
    \State Compute $S_{i,\text{link}}(t)$ according to \refequ{width}.
    
    % \State $\text{sending\_flow} \gets c \cdot \text{num\_peds}[t] + (1 - c) \cdot S_{i,\mathrm{boundary}}(t)$

    \If{$k(t) \le k_c$} \Comment{free-flow regime}
        \State $\textcolor{purple}{S_{i,\text{stochastic}}(t)} \gets \texttt{crowd\_diffusion}(\tau(t),\gamma)$
        \Else \Comment{congested regime}
        \State Compute $p_{\text{rel}}$ with \refequ{eq:release_probability}.
        \State $\textcolor{purple}{S_{i,\text{stochastic}}(t)} \sim \mathrm{Bin}\left(\lfloor S_{i,\text{boundary}}(t) \rfloor, p_{\text{rel}}\right)$

    \EndIf
    \If{$p_{\text{activity}} > 0$} \Comment{activity effect}
        \State $X_i(t) \sim \mathrm{Bin}\left(\lfloor S_{i,\text{stochastic}}(t) \rfloor, p_{\text{activity}}\right)$
        \State $\textcolor{purple}{S_{i,\text{stochastic}}(t)} \gets S_{i,\text{stochastic}}(t) - X_i(t)$
    \EndIf

    \State Update $S_i(t)$ with \refequ{eq:stochastic}.
    % \State Smoothing the sending flow with last step: \State $S_i(t) \gets \delta \cdot S_i(t) + (1-\delta) \cdot S_i(t-1)$
    \EndProcedure
\Ensure $S_i(t)$
\end{algorithmic}
\end{algorithm}

\section{Route Choice and Node Flow Assignment}\label{route_node}
After the sending and receiving flows of each link have been determined, two further steps are required at every node. First, the route choice model determines how the flow on each upstream link should be divided among the feasible downstream links. Second, the node model converts these desired turning fractions into feasible movement flows under the sending and receiving constraints. This section presents these two connected components.

\subsection{Dynamic Route Choice with Utility}\label{utility}
Route choice must respond to time-varying congestion and control actions. We therefore introduce a dynamic utility-based route-choice model, following~\cite{daamen2004modelling}. At each simulation step, the model evaluates the feasible downstream links for every origin--destination (OD) pair, converts their attributes into aggregate utility scores, and maps these scores to turning fractions.

For each OD pair, we first generate $k$ shortest candidate paths. At a node, a downstream link is feasible if it belongs to at least one candidate path for that OD pair. Each feasible link is associated with a configurable attribute vector, which may include remaining distance, current travel time, comfort, congestion, and control attributes. Users can also include scenario-specific attributes, such as activity spaces or greenery.

Here, ``utility'' denotes an aggregate score used to rank feasible links and allocate the flow of each OD pair. It is not a behavioral utility sampled for each pedestrian. We denote the utility score of link $\ell$ for OD pair $\pi$ at time $t$ by $\mathcal{U}_{\ell}^{\pi}(t)$:
\begin{equation}
\mathcal{U}_{\ell}^{\pi}(t)
= \boldsymbol{\theta}^{\mathsf{T}}\mathbf{x}_{\ell}^{\pi}(t) + \eta_{\ell}(t).
\label{eq:aggregate_utility}
\end{equation}

The vector $\mathbf{x}_{\ell}^{\pi}(t)$ contains the observed link attributes that are dependent on $\pi$, and $\boldsymbol{\theta}$ contains their coefficients. The signs and magnitudes of these coefficients determine how each attribute affects the allocation. For example, distance and travel time normally have negative coefficients, whereas comfort has a positive coefficient. The term $\eta_{\ell}(t)\sim\mathcal{N}(0,\sigma^2)$ is a temporary shock shared at the link level. It captures conditions absent from the observed attributes, such as a temporary obstruction or a change in collective perception. At time $t$, the shock changes the utility score for all aggregate flow considering link $\ell$.

Let $\ell_{\uparrow}$ denote the upstream link, $\ell_{\downarrow}$ a feasible downstream link, and $\pi$ an OD pair in $\mathcal{P}_{od}$. We map the utility scores to downstream link choice for each OD pair using a softmax allocation rule:
\begin{equation}
P_{t}(\ell_{\downarrow} \mid \ell_{\uparrow}, \pi)
= \frac{\exp\big(\mathcal{U}_{\ell_{\downarrow}}^{\pi}(t)\big)}
{\sum_{\ell' \in \mathcal{D}^{\pi}(\ell_{\uparrow})} \exp\big(\mathcal{U}_{\ell'}^{\pi}(t)\big)},
\label{eq:od_turning_fraction}
\end{equation}
where $\mathcal{D}^{\pi}(\ell_{\uparrow})$ denotes the downstream links from $\ell_{\uparrow}$ that are feasible for OD pair $\pi$. Although written as a probability, $P_t(\ell_{\downarrow}\mid\ell_{\uparrow},\pi)$ is implemented directly as the fraction of the macroscopic flow for OD pair $\pi$ assigned to $\ell_{\downarrow}$. The model does not generate individual choices. A link with a higher utility score receives a larger share of the flow, and the shares over all feasible downstream links sum to one.

Because several OD pairs can be present on the same upstream link, we combine their fractions to obtain the turning fraction used by the node model:
\begin{equation}
P_{t}(\ell_{\downarrow}\mid \ell_{\uparrow})
= \sum_{\pi \in \mathcal{P}_{od}}
\underbrace{P_{t}(\ell_{\downarrow}\mid \ell_{\uparrow}, \pi)}_{\text{link choice given OD}}
\underbrace{P_{t}(\pi\mid \ell_{\uparrow})}_{\text{OD ratio}}.
\label{eq:turning-od}
\end{equation}
Here, $P_{t}(\pi \mid \ell_{\uparrow})$ is the proportion of flow on $\ell_{\uparrow}$ associated with OD pair $\pi$. It is calculated from the OD demand at time $t$. Equation~\ref{eq:turning-od} therefore combines the utility allocations for all OD pairs into one turning fraction for each movement from an upstream link to a downstream link.

\begin{figure}[htb!]
    \centering
    \includegraphics[width=\columnwidth]{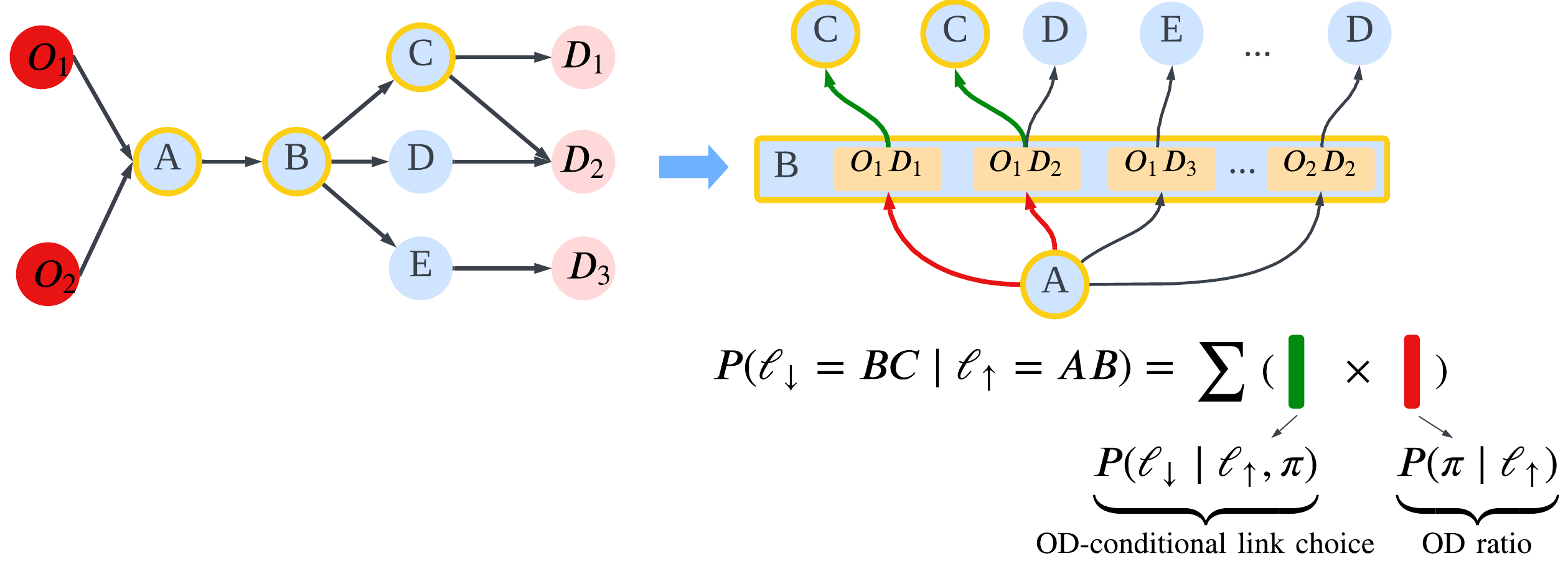}
    \caption{Route choice using utility at node B. The utility scores for each OD pair determine the fractions assigned to downstream links. These fractions are then weighted by the OD flow proportions on each upstream link.}
    \label{route}
\end{figure}

The computation is illustrated in \reffig{route}. The left panel shows a network with two origins and three destinations. Node B lies on the candidate paths of six OD pairs. To determine the turning fraction from link $\textit{A--B}$ to link $\textit{B--C}$, we first identify the OD pairs for which this movement is feasible. We then compute their fractions from utility using \refequ{eq:od_turning_fraction} and weight them by the corresponding OD proportions using \refequ{eq:turning-od}.

\subsection{Flow Assignment at Nodes}\label{node_model}
The route choice model provides the turning fraction $p^n_{ij}$ from upstream link $i$ to downstream link $j$ at node $n$. The node model then determines the realized movement flow $q^n_{ij}$ from the sending and receiving flows. For the simulations in this paper, we use the direct allocation rule from~\cite{yperman2005link}:

\begin{equation}
    q^{n}_{ij} = \min\left(\frac{p^{n}_{ij}S^{n}_i}{\sum_{l\in \mathcal{I}_n}p^{n}_{lj}S^{n}_l}R^{n}_j,\,p^{n}_{ij}S^{n}_i \right),
    \label{classic}
\end{equation}
where $\mathcal{I}_n$ is the set of incoming links at node $n$, $S_i^n$ is the sending flow of link $i$, and $R_j^n$ is the receiving flow of link $j$. The rule distributes the available downstream space according to the turning fractions and upstream demand. \emph{PedNStream} also provides an optional flow maximization formulation for evacuation scenarios~\cite{lilasathapornkit2022dynamic}. It maximizes node throughput subject to the sending, receiving, and turning fraction constraints and can be selected through the configuration file.

\section{Simulation Framework and Control Integration}\label{architecture}
\begin{figure*}[!ht]
    \centering
    \includegraphics[width=\textwidth]{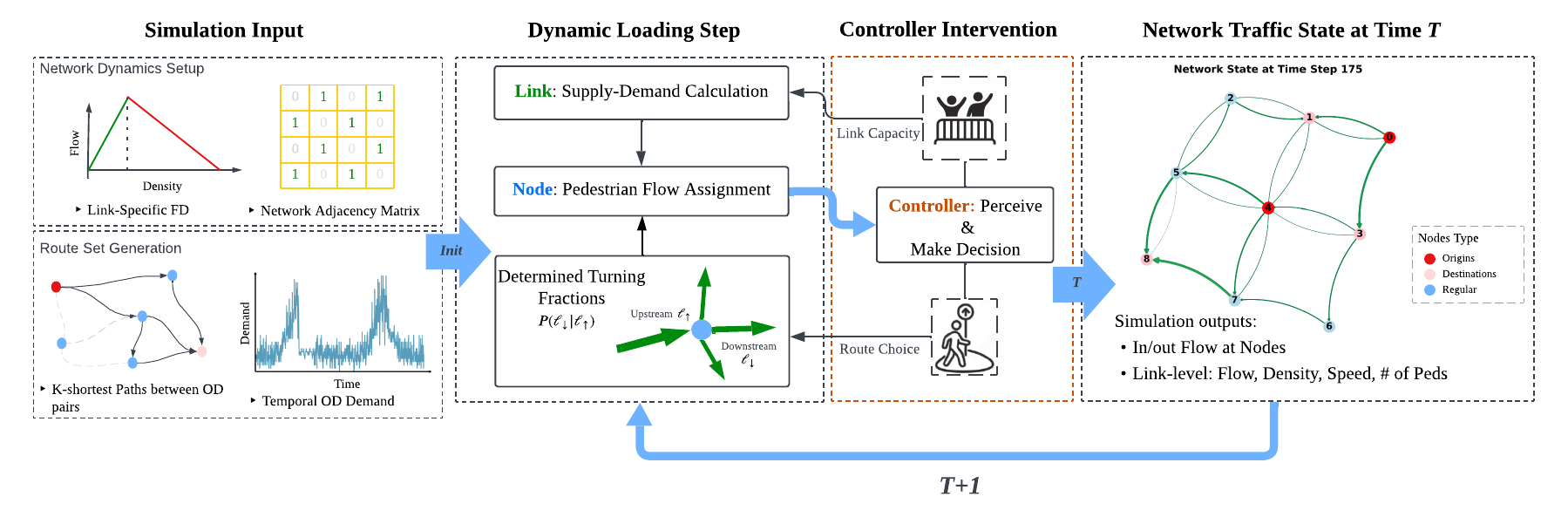}
    \caption{Simulation framework of \emph{PedNStream}. The controller observes the updated network state and modifies the control variables used in the next loading step.}
    \label{overall}
\end{figure*}

\emph{PedNStream} combines the link dynamics, route choice, node flow assignment, and controller interfaces in a repeated simulation pipeline. \reffig{overall} shows how these components exchange information during a simulation. The main text presents this conceptual framework and its control interface. Details of the software structure and scenario configuration are provided in Appendix~\ref{software_architecture}.

\subsection{Overall Simulation Framework}
As shown in \reffig{overall}, the pipeline begins with the \textit{Simulation Input} phase. A scenario specifies the network dynamics, including link-specific fundamental diagrams (FDs) and the network adjacency matrix. It also defines the route set through temporal OD demand and the $K$ shortest paths between OD pairs. After initialization, \emph{PedNStream} repeats the dynamic loading process over $t=1,\ldots,T_{\max}$. At each time step, the route choice module updates turning fractions, the node model assigns flows between connected links, and the link model updates flow, density, and walking speed. At each control interval, the controller receives the updated network state and uses selected state variables to adjust route guidance or physical capacities. These actions take effect in the next loading step, closing the feedback loop. The same workflow can therefore be used to evaluate different traffic models and control strategies.

\begin{figure}[!htb]
    \centering
    \includegraphics[width=.45\textwidth]{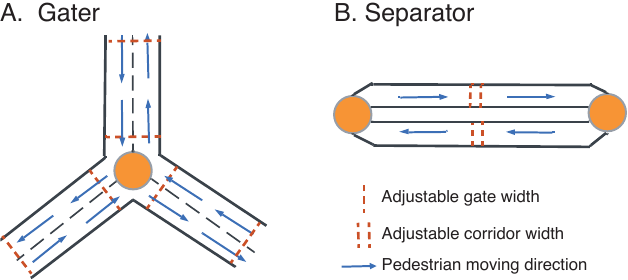}
    \caption{Control devices in \emph{PedNStream}. (A) A gater changes the available width at a link entrance or exit to regulate flow through a node. (B) A separator allocates the available street width between the two walking directions and reduces bidirectional interference.}
    \label{control}
\end{figure}
\subsection{Controller Layer and Control Device Design}\label{controller}
The controller layer provides a common observation and action interface for closed-loop experiments. At each control interval, a controller receives selected network states, such as link densities, flows, or travel times. It then returns an action that changes a physical or operational parameter of the network. The action is clipped to its feasible range and applied before the next network loading update. This interface supports rule-based, optimization-based, and learning-based controllers without embedding a specific control method in the pedestrian flow model.

% Many studies prescribe a target inflow or an exact number of pedestrians allowed to enter an area~\cite{liang2023reinforcement,10944426}. Such traffic variables cannot always be enforced directly in open pedestrian environments. 

In \emph{PedNStream}, controllers act by adjusting physical device settings, especially the available passage width~\cite{molyneaux2021design}. We implement two complementary device abstractions, the \textit{Gater} and the \textit{Separator}, as illustrated in \reffig{control}. They represent two common forms of operational intervention. A Gater regulates movement through an intersection or access point, while a Separator reallocates shared street space between walking directions along a link. Together, they allow the framework to test both flow regulation and directional space allocation without prescribing an exact pedestrian flow.

The two devices are integrated directly into the model's link structure, where each walking direction is represented as a distinct unidirectional link, allowing opposing directions to share the same physical street space. A \textit{Gater} operates at the entrance or exit of a link, dynamically adjusting its width to modify the sending or receiving capacity according to \refequ{width}, which indirectly influences route allocation. Conversely, a \textit{Separator} adjusts the width split between two opposing links, thereby altering their effective operational areas and mitigating bidirectional interference. To demonstrate this interface, we implement both rule-based and pressure-based gate controllers as interpretable examples; both variants observe local densities to output target gate widths, with their detailed formulations provided in Appendix~\ref{apd}.

\section{Synthetic Scenario Evaluation}\label{syn_exp}
We first evaluate \emph{PedNStream} on four simple synthetic networks. Each scenario focuses on a specific flow pattern: interaction between opposite flows, queue formation and spillback, recovery after a demand surge, or route changes under congestion. Examining these patterns separately helps determine whether the simulator produces the expected behavior before it is applied to larger networks. The four scenarios are summarized in Table~\ref{tab:networks}.
\subsection{Scenario and Network Design}
\begin{figure}[!htb]
    \centering
    \includegraphics[width=\columnwidth]{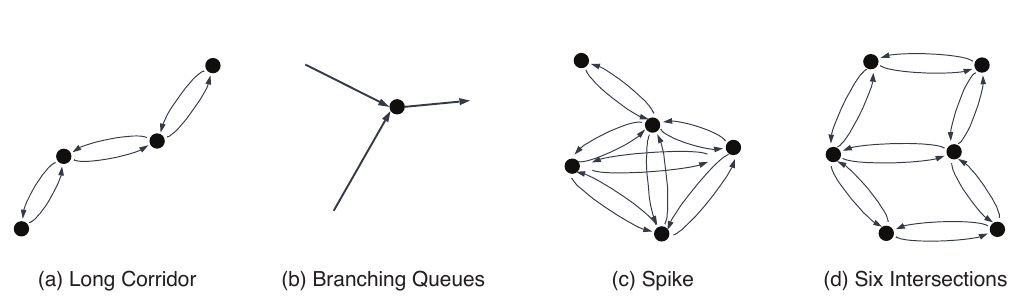}
    \caption{Four synthetic networks.}
    \label{syn}
\end{figure}

\begin{figure*}
    \centering
    \includegraphics[width=\textwidth]{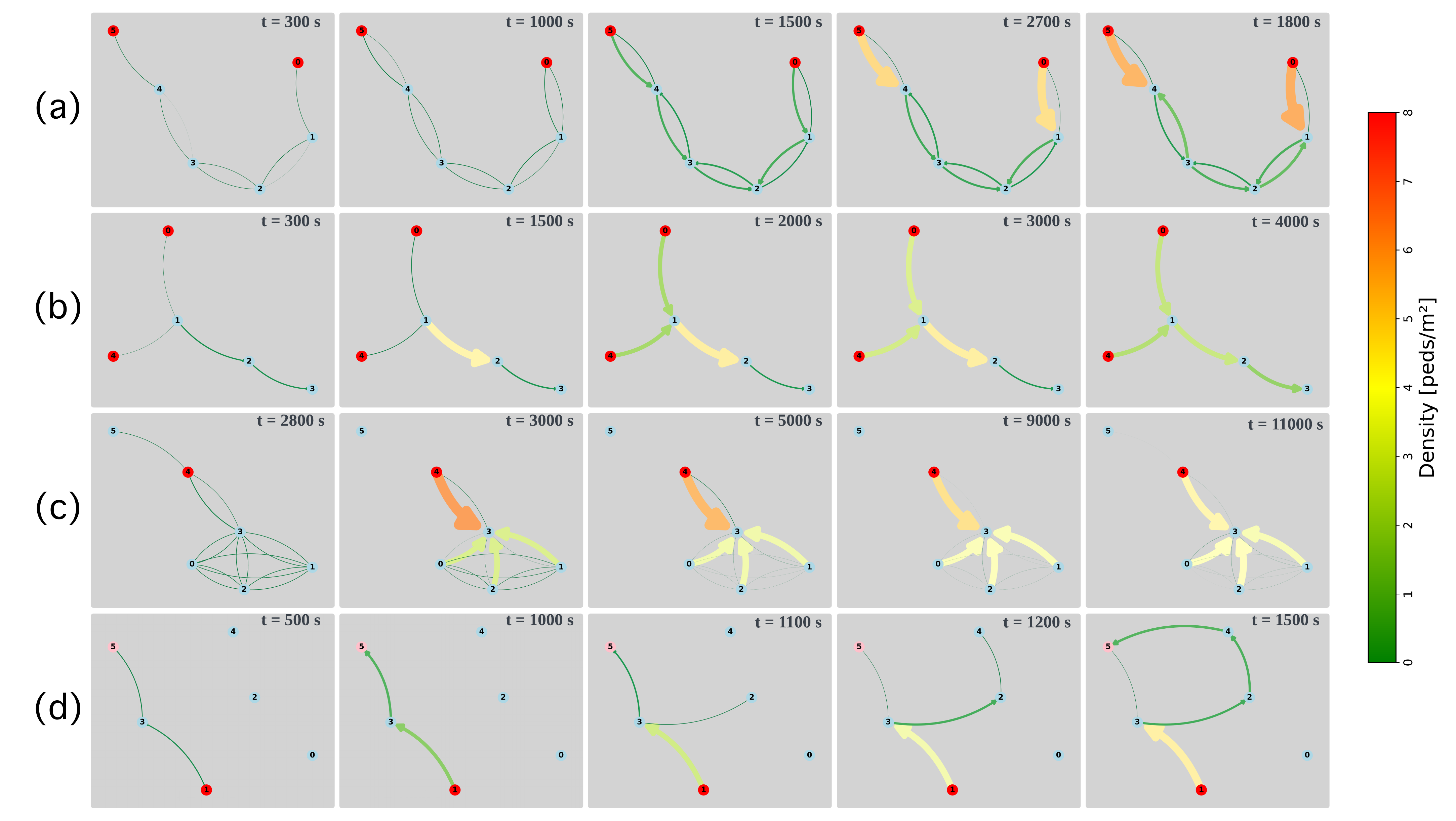}
    % \Description[ ]
    \caption{Simulation snapshots for four synthetic networks. (a) Long corridor. (b) Branching queues. (c) Spike. (d) Six intersections.}
    \label{vis}
\end{figure*}

The synthetic network layouts are shown in Fig.~\ref{syn}, and representative simulation snapshots are shown in \reffig{vis}. In all cases, red nodes denote origins and pink nodes denote destinations. For scenarios with OD pairs, \emph{PedNStream} precomputes multiple shortest paths to enable dynamic route choice during simulation.

\begin{table}[htbp]
\centering
\caption{Synthetic network designs and scenario focus}
\label{tab:networks}
\renewcommand{\arraystretch}{1.2} % Improves row spacing
\begin{tabular}{>{\raggedright\arraybackslash}m{3cm} >{\raggedright\arraybackslash}m{5cm}}
\hline
\textbf{Network Design} & \textbf{Scenario Focus} \\
\hline
Long Corridor~(Fig.~\ref{vis}(a)) & \textit{Pedestrians enter from both ends to evaluate bidirectional flow interactions.} \\
Branching Queues~(Fig.~\ref{vis}(b)) & \textit{A downstream bottleneck is introduced to examine queue formation and spillback.} \\
Spike~(Fig.~\ref{vis}(c)) & \textit{A sudden demand surge causes intersection blockage, testing congestion recovery.} \\
Six Intersections~(Fig.~\ref{vis}(d)) & \textit{Dynamic route choice is evaluated under time-varying traffic conditions.} \\

\hline
\end{tabular}
\end{table}
% \subsection{Results Visualization}

% \subsection*{Long Corridor}
% \noindent\textbf{Long Corridor:} We simulate bidirectional pedestrian flow in a long corridor to examine counterflow interference. As shown in \reffig{vis}(a), pedestrians enter from both ends and initially move under near free-flow conditions. As demand increases, the two opposing streams compete for the same corridor space, reducing walking speed and causing congestion to emerge and propagate toward both entrances. This scenario highlights the model's ability to capture mutual hindrance and queue formation in bidirectional pedestrian flow.

\noindent\textbf{Long Corridor:} As shown in \reffig{vis}(a), the two opposing streams initially move under near free-flow conditions. As demand increases, they compete for the same corridor space, reducing walking speeds and causing congestion to form and propagate toward both entrances. This result shows that the model captures interference and queue formation in bidirectional pedestrian flow.

% We then compare two control settings, one with a flow separator and one without, using identical demand patterns. Figure~\ref{separator} shows that separators significantly mitigate congestion by reducing bidirectional interference.
% \begin{figure}[!htp]
% \centering
% \includegraphics[width=\columnwidth]{fig/long.pdf}
% \caption{Bidirectional flow in a long corridor ($k_{\mathrm{critical}}=2$, $k_{\mathrm{jam}}=6$). (a) Without separators. (b) With separators.}
% \label{separator}
% \end{figure}

\noindent\textbf{Branching Queues:} With the back gate of link \textit{2--3} narrowed to 0.5~m, pedestrians accumulate on link \textit{1--2} between 300~s and 3000~s, as shown in \reffig{vis}(b). The resulting queue propagates upstream into links \textit{0--1} and \textit{4--1}. When the gate is reopened at 3000~s, the queue gradually dissipates and the network returns toward normal flow conditions. These results show how a local capacity restriction can cause queue growth and spillback through the upstream network.
% \noindent\textbf{Branching Queues:} In this network (\reffig{vis}(b)), pedestrians enter from both ends of a fork-shaped corridor. A downstream bottleneck is introduced by narrowing the back gate of link \textit{2--3} to 0.5 m. As a result, pedestrians begin to accumulate on link \textit{1--2} between 300 s and 3000 s, and the resulting queue propagates upstream into links \textit{0--1} and \textit{4--1}. This scenario illustrates how a local capacity restriction can trigger queue growth and spillback through the upstream network. At 3000 s, the back gate is reopened to 1.0 m, after which the queue gradually dissipates and the network returns toward normal flow conditions.

% \subsection*{Spike}
\noindent\textbf{Spike:} After the demand surge, pedestrians become blocked around intersection \textit{3}, and near-gridlock forms at \(t=5000\)~s in \reffig{vis}(c). Congestion then gradually eases because stochasticity in the sending flow creates additional movement opportunities. This result represents the assumption that pedestrians can occasionally move through local gaps even under severe congestion.

\noindent\textbf{Six Intersections:} For pedestrians travelling from node \textit{1} to node \textit{5}, the route utility combines distance, link density, and capacity. Initially, they choose the shortest path through link \textit{1--3}. After the front gate of link \textit{1--3} is narrowed to 0.1~m at 1100~s, a larger share reroutes toward link \textit{3--2}. These flow shifts are consistent with the route-utility formulation in \refsec{utility}.
% \noindent\textbf{Six Intersections:} This scenario examines dynamic route choice using the utility function in \refsec{utility}, which incorporates distance, link density, and capacity. Initially, pedestrians traveling from node \textit{1} to node \textit{5} choose the shortest path via link \textit{1--3}. At 500 s, increasing density on link \textit{1--3} prompts a small fraction of pedestrians to divert via link \textit{1--0}. At 1100 s, the front gate of link \textit{1--3} is narrowed to 0.1 m, triggering substantial rerouting toward link \textit{3--2}. The resulting flow shifts are consistent with the utility design, supporting the validity of the route choice mechanism.

Overall, these experiments show that \emph{PedNStream} can reproduce key pedestrian traffic phenomena, including bidirectional interference, queue spillback, congestion recovery, and adaptive routing, while responding realistically to control interventions. The simulator's flexibility supports systematic testing of both static network designs and dynamic control strategies under various conditions.

\subsection{Pedestrian Flow Pattern Comparison}
\begin{figure}[!htb]
    \centering
    \includegraphics[width=\columnwidth]{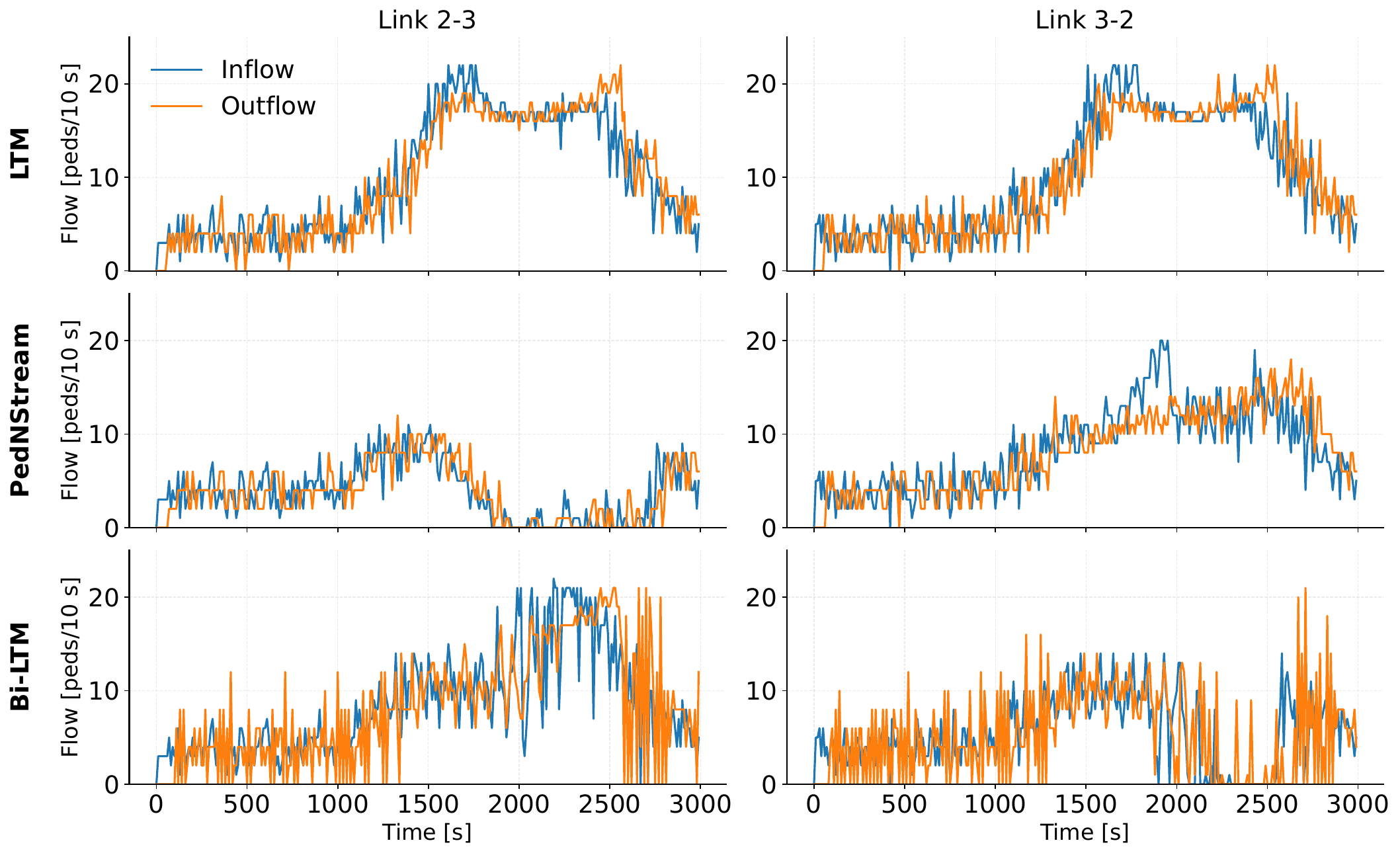}
    \caption{Qualitative comparison of inflow and outflow patterns on link pair \textit{2--3}/\textit{3--2} in the Long Corridor scenario. From top to bottom: LTM, \emph{PedNStream}, and bidirectional LTM.}
    \label{long_corridor_flow}
\end{figure}

To examine the effect of the proposed link dynamics, we compare the inflow and outflow patterns generated by three models: the original LTM, \emph{PedNStream}, and the bidirectional LTM. We select the Long Corridor and Spike networks because they isolate two key behaviors. The Long Corridor tests interactions between opposing pedestrian streams, whereas the Spike network tests flow behavior under near-gridlock conditions. Figure~\ref{long_corridor_flow} presents a representative link pair in the Long Corridor scenario. The left and right columns correspond to the two opposing travel directions. In the original LTM, the two directions exhibit similar flow profiles because the model does not explicitly account for interactions between opposing streams. In contrast, both \emph{PedNStream} and the bidirectional LTM show direction-dependent flow patterns, where the outflow in one direction is influenced by the inflow in the opposite direction. Compared with the bidirectional LTM, \emph{PedNStream} produces smoother and less oscillatory flow patterns because of the added stochastic and diffusion effects.

\begin{figure}[!htb]
    \centering
    \includegraphics[width=\columnwidth]{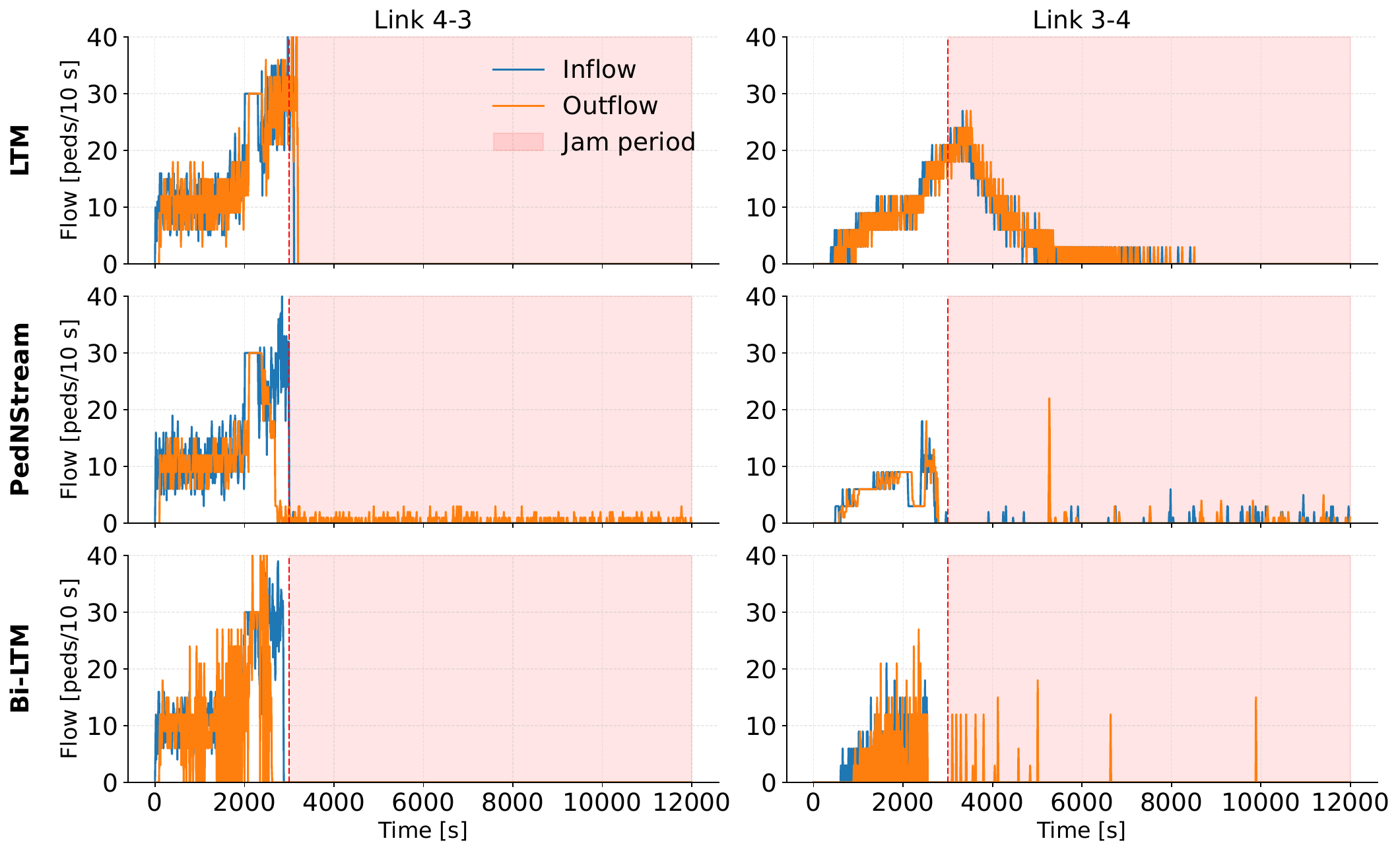}
    \caption{Qualitative comparison of inflow and outflow patterns on link pair \textit{4--3}/\textit{3--4} in the Spike scenario under near-gridlock conditions. From top to bottom: LTM, \emph{PedNStream}, and bidirectional LTM.}
    \label{spike_flow}
\end{figure}

Figure~\ref{spike_flow} provides a second comparison in the Spike network under near-gridlock conditions. In this case, link \textit{4--3} reaches jam density after approximately 2800~s (see \reffig{vis}(c)), after which the demand on that link drops to zero. Accordingly, the inflow of link \textit{4--3} also drops to zero, as shown in the left column of \reffig{spike_flow}. The standard LTM still shows relatively high inflow and outflow on link \textit{3--4}, which carries pedestrians in the opposite direction. This implies unrealistically unconstrained movement under severe congestion. By contrast, the bidirectional LTM almost completely suppresses the inflow on link \textit{3--4}, indicating that no usable space remains. \emph{PedNStream} produces an intermediate behavior: flow on link \textit{3--4} is strongly reduced, but a small releasing flow is still maintained. This is more consistent with the intended modeling assumption that pedestrians can occasionally squeeze through local gaps even under highly congested conditions.

Together, these examples qualitatively show that \emph{PedNStream} preserves the bidirectional interaction mechanism while producing smoother and more dispersed flow profiles because of the added stochastic and diffusion effects.

\section{Real Network Evaluation} \label{real_network}
After the mechanism-level and comparative tests above, we examine \emph{PedNStream} on two real-world sidewalk networks (Delft and Melbourne city centers) to assess whether the same model scales to larger, more heterogeneous settings and remains broadly consistent with observed pedestrian counts.

\subsection{Large-Scale Dynamics in Delft}
\begin{figure*}[htp!]
\centering
\includegraphics[width=1\textwidth]{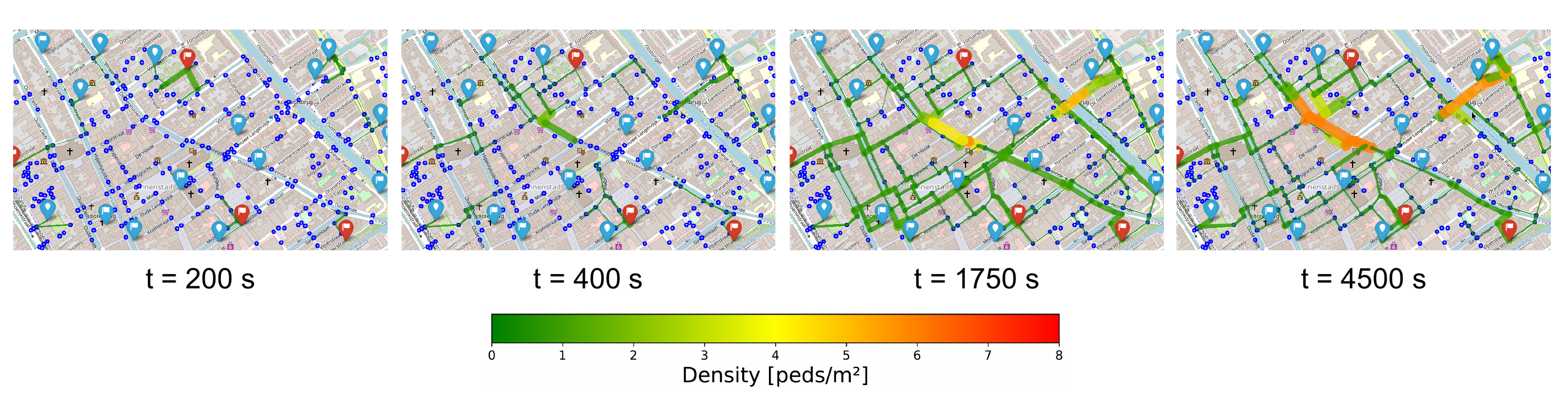}
\caption{Simulation visualization of Delft city centre. Red markers denote nodes serving as both origins and destinations, blue markers with flags indicate destinations only, and normal markers represent origins. Remaining circular points depict link connection nodes.}
\label{delft}
\end{figure*}

The Delft city centre network consists of 298 nodes and 818 links. We simulate 500 steps (84 minutes) with a total demand of 46,501 pedestrians. The resulting network state is shown in \reffig{delft}.
For each OD pair, three paths are precomputed to generate the routes set. The simulation result shows that \emph{PedNStream} can represent network-wide pedestrian propagation and the emergence of localized congestion. After approximately 4,500 s, two main hotspots appear: the central bridge and the main road connecting the major origins and destinations. These locations serve as shared critical corridors where heavy demand concentrates and congestion is most visible.

\begin{figure}[htbp]
    \centering
    \subfloat[Representative paths \textit{0--8} (light yellow) and \textit{142--80} (light blue) in the Delft network.\label{delft_paths}]{%
        \includegraphics[width=.9\columnwidth]{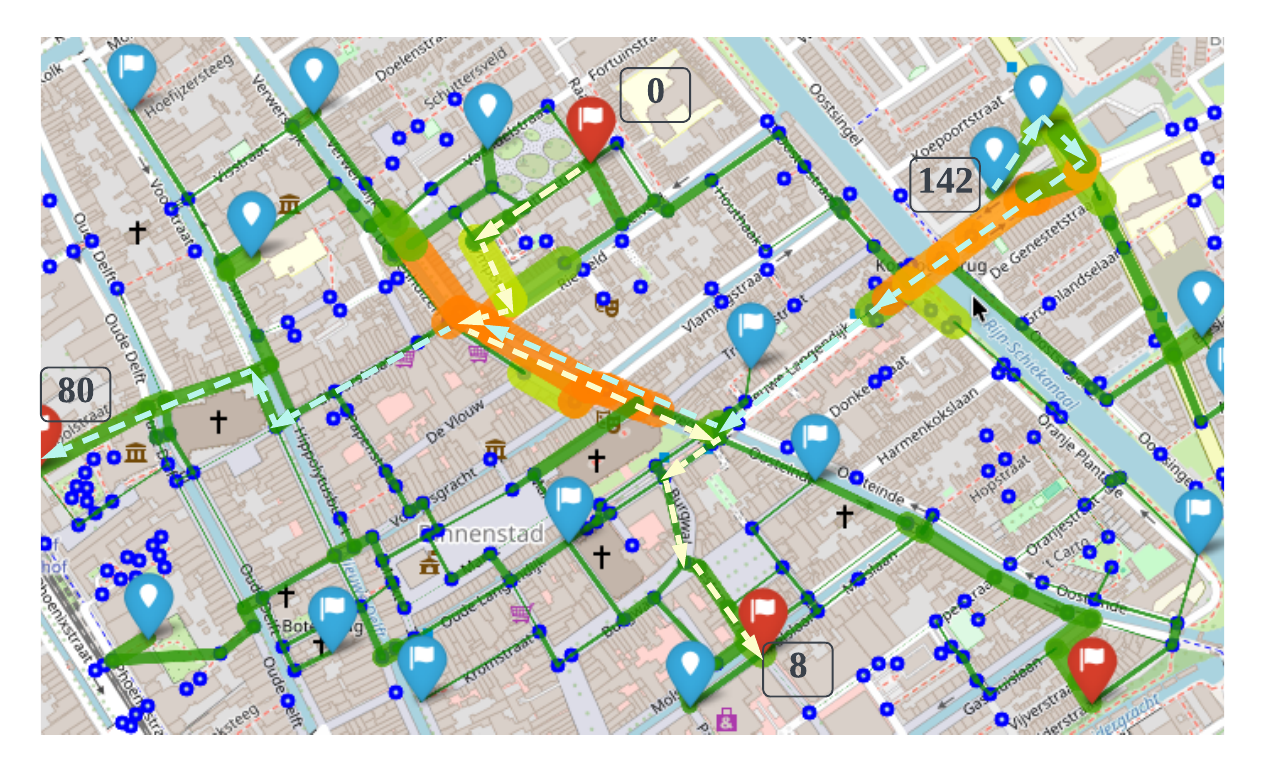}%
    }

    \medskip

    \subfloat[Space-time diagram for path \textit{0--8}.\label{0-8}]{%
        \includegraphics[width=.75\columnwidth]{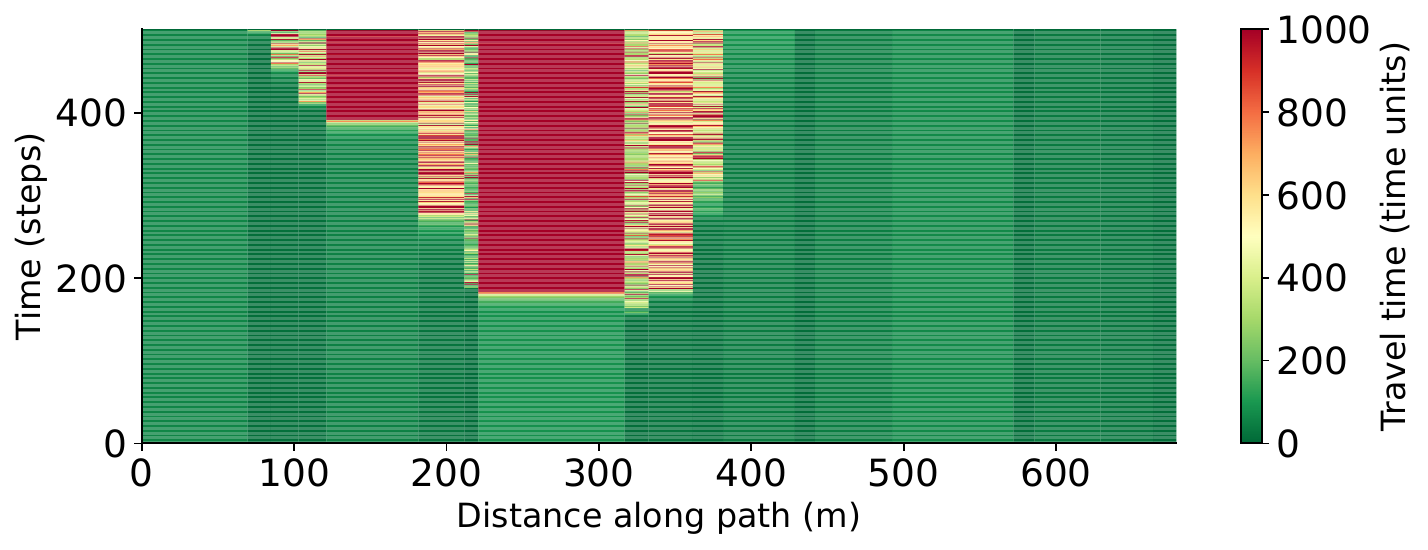}%
    }

    \medskip

    \subfloat[Space-time diagram for path \textit{142--80}.\label{142-80}]{%
        \includegraphics[width=.8\columnwidth]{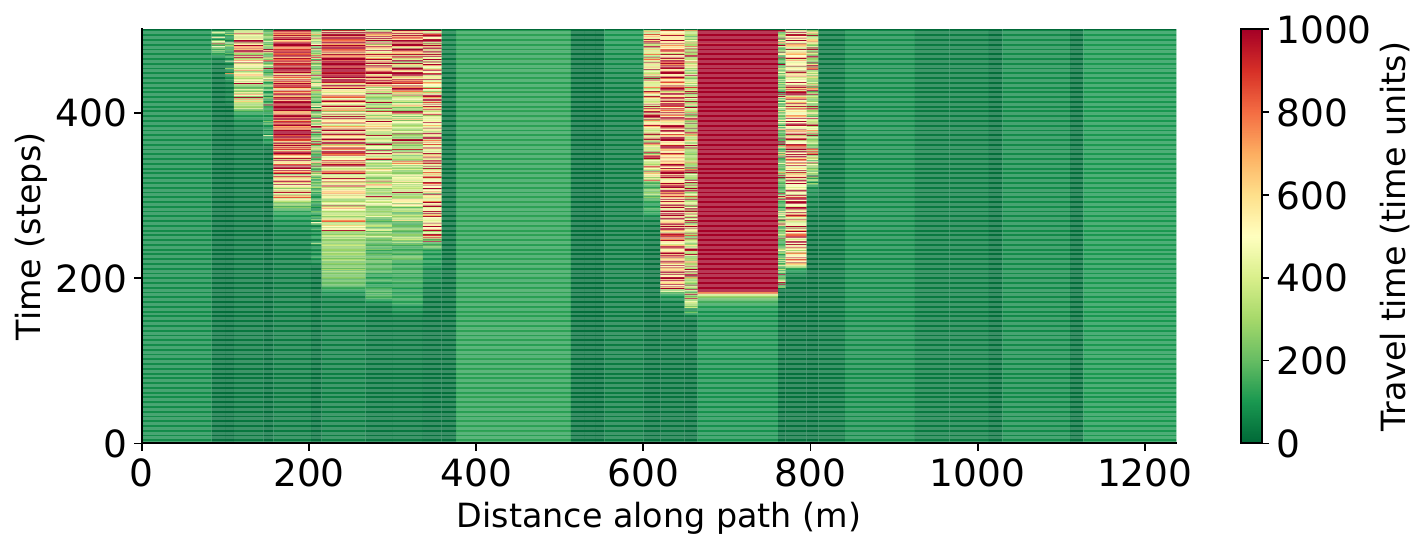}%
    }
    \caption{Space-time diagrams of two main paths in the Delft center.}
    \label{fig:vertical}
\end{figure}

To examine the simulated traffic dynamics in more detail, we analyze two representative OD paths, \textit{0--8} (north to south) and \textit{142--80} (east to west), shown in \reffig{delft_paths}. The space-time diagram for path \textit{0--8} in \reffig{0-8} shows that severe congestion first emerges around 200--300 m along the path and then gradually propagates upstream. This pattern is expected when a downstream restriction, such as a narrow bridge or conflicting movements at an intersection, discharges pedestrians more slowly than they arrive. Pedestrians then accumulate upstream, and the tail of the queue moves backward along the path. The second diagram, \reffig{142-80}, shows two wedge-shaped congestion patterns that correspond to the two main hotspots highlighted in \reffig{delft_paths}. Together, these path-based views provide a more detailed picture of how congestion forms, propagates, and interacts with route structure in the Delft network.

\subsection{Partial Validation with Melbourne Count Data}
\begin{figure}[htbp]
  \centering
  \subfloat[Melbourne city-centre sidewalk network. Blue marks denote the location of the sensors.\label{mel}]{%
    \includegraphics[width=.8\columnwidth]{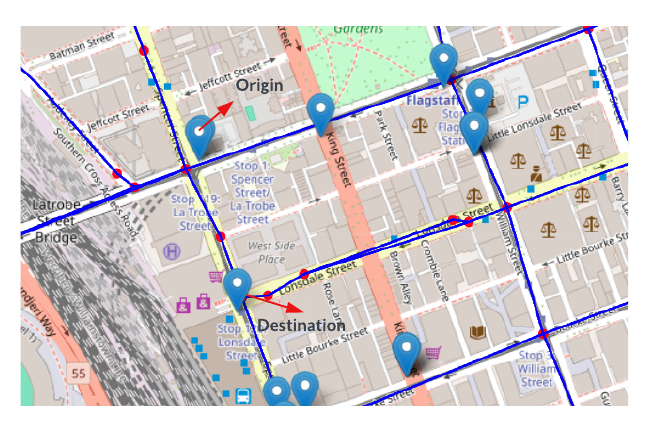}%
  }

  \medskip

  \subfloat[Observed and simulated flow comparison at a representative location. In this comparison, $k_c=2\,\mathrm{peds/m^2}$ and $v_f=1.1\,\mathrm{m/s}$. The diffusion coefficient $\gamma$ is set to 0.1, and the activity probability $p_{\text{activity}}$ is 0.5.\label{fig:sim}]{%
    \includegraphics[width=\columnwidth]{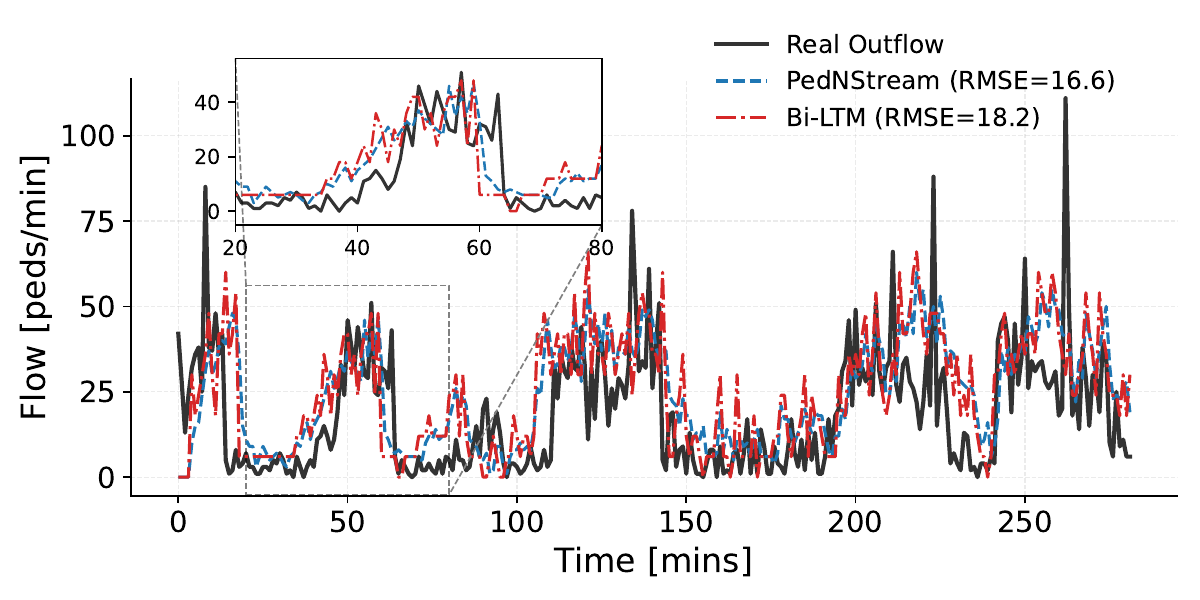}%
  }
  \caption{Comparison between real counting data and simulated data.}
  \label{fig:melbourne_comparison}
\end{figure}

After the synthetic tests, we use count data from the Melbourne pedestrian counting system \footnote{\href{https://data.melbourne.vic.gov.au/explore/dataset/pedestrian-counting-system-sensor-locations}{Melbourne pedestrian counting system sensor locations dataset}.} to assess whether \emph{PedNStream} produces realistic aggregate flow patterns in an urban network. The data contain counts but no OD demand or turning-fraction information, so this is a partial validation rather than a strict predictive test. \reffig{mel} shows the sensor locations (blue markers) and simulated network nodes (red points). At a representative location in \reffig{fig:sim}, both \emph{PedNStream} and the bidirectional LTM follow the observed count trend. \emph{PedNStream} has a lower root mean square error (RMSE = 16.6) than Bi-LTM (RMSE = 18.2), suggesting that its stochastic link dynamics capture local fluctuations more closely.

% We then evaluate the model at the network level using all 92 sensors. In each run, we select 5, 10, or 20 sensors as observed inputs and use their count series to drive the simulation. The remaining sensors are held out for evaluation. For each held-out sensor and each 15~min period, we compare the simulated count with the observed count. Two GEH matching rates report the percentage of these comparisons with a GEH value below 5 or below 10. A GEH value below 5 is commonly interpreted as good agreement \cite{dowling2004traffic}. We also report the volume ratio, NRMSE, and NDTW. These measures describe agreement in total flow, the size of count differences, and the timing and shape of the flow pattern, respectively. \reftbl{tab:validation_results} reports the mean and standard deviation across seven runs.
% Because ground-truth OD flows and turning fractions are unavailable, this network-level experiment evaluates whether \emph{PedNStream} can reproduce held-out aggregate counts from partial observations; it does not test prediction at the OD or route level.

We evaluate the model using 92 Melbourne sensors. For each observation budget
($m\in\{5,10,20\}$), we conduct seven runs using independent observed-sensor sets, generated with the fixed random seeds. In each run, the $m$ selected sensors are the only sensors whose count series are supplied to the simulation. Each selected sensor is
mapped to its associated network node and its directional count series is
aggregated to 15-minute intervals and used as the corresponding directional
boundary inflow. The
remaining $92-m$ sensors are never used as simulation inputs and form the
spatial holdout set.

For each held-out sensor, we compare the simulated and observed counts in every 15-minute interval. We evaluate the results using GEH~\cite{dowling2004traffic}, volume ratio, normalized root mean square error (NRMSE), and normalized dynamic time-warping (NDTW) distance. The volume ratio measures differences in total flow, NRMSE measures count error, and NDTW measures differences in the timing and shape of flow patterns. \reftbl{tab:validation_results} reports the mean and standard deviation across seven runs.

We use 24 hours of minute-level data, from 15:55 on May~24, 2025 to 15:55 on
May~25, 2025. To reduce the influence of noisy or discontinuous series, we
aggregate all counts into 15-minute intervals. The holdout is spatial rather
than temporal: observed-sensor series may be used over the full 24-hour
window to drive the simulation, but held-out sensor series are used only for
evaluation. The spatial KNN baseline is evaluated using exactly the same
observed and held-out sensor sets; it estimates each held-out sensor from a
distance-weighted average of its $k$ nearest observed sensors.

For the OD-informed \emph{PedNStream} variant, the OD weights are estimated
from the observed sensors in each run. We compute Pearson correlations over
the calibration portion of the 24-hour window, set negative correlations to
zero, and normalize the remaining values to obtain the OD ratios in
\refequ{eq:turning-od}. Held-out sensor series are not used for calibration.
This experiment therefore compares \emph{PedNStream} with spatial KNN under
partial spatial observations. It shows their relative reconstruction
performance, but does not establish predictive realism or recover the true OD
demand and turning fractions.
\begin{table*}[t]
\centering
\caption{Melbourne partial-observation validation. \textbf{Obs.} denotes the number of sensors treated as observed inputs, and metrics are computed on the remaining sensors. Entries report mean (std) across 7 runs. \textbf{Bold} indicates the best value per setting. $\uparrow$ higher is better; $\downarrow$ lower is better; $\approx$1 closer to 1 is better. GEH values below 5 are
commonly interpreted as good agreement \cite{dowling2004traffic}.}
\label{tab:validation_results}
\small
\setlength{\tabcolsep}{8pt} % Increased spacing for readability
\begin{tabular}{ll ccccc}
\toprule
\textbf{Obs.} & \textbf{Method} 
  & \textbf{GEH < 5 (\%)} $\uparrow$ 
  & \textbf{GEH < 10 (\%)} $\uparrow$ 
  & \textbf{Vol. Ratio} $\approx$1 
  & \textbf{NRMSE} $\downarrow$ 
  & \textbf{NDTW} $\downarrow$ \\
\midrule

\multirow{3}{*}{5} 
  & PedNStream (w/o OD) & \textbf{6.57 \scriptsize(4.58)} & \textbf{13.71 \scriptsize(5.44)} & 1.23 \scriptsize(0.57) & \textbf{1.60 \scriptsize(0.45)} & \textbf{1.25 \scriptsize(0.37)} \\
  & PedNStream (w/ OD)  & 4.43 \scriptsize(3.41)          & 10.86 \scriptsize(4.85)          & \textbf{1.16 \scriptsize(0.44)} & 1.63 \scriptsize(0.37)          & 1.32 \scriptsize(0.31)          \\
  & KNN      & 2.71 \scriptsize(2.36)          & 3.71 \scriptsize(2.98)           & 5.06 \scriptsize(1.79)          & 4.57 \scriptsize(1.59)          & 4.47 \scriptsize(1.60)          \\
\cmidrule(lr){1-7}

\multirow{3}{*}{10} 
  & PedNStream (w/o OD) & 6.00 \scriptsize(2.96)          & 11.56 \scriptsize(2.92)          & 2.33 \scriptsize(0.82)          & \textbf{2.51 \scriptsize(0.77)} & \textbf{2.05 \scriptsize(0.69)} \\
  & PedNStream (w/ OD)  & \textbf{6.22 \scriptsize(3.87)} & \textbf{13.00 \scriptsize(4.50)} & \textbf{2.31 \scriptsize(0.71)} & 2.56 \scriptsize(0.64)          & 2.13 \scriptsize(0.57)          \\
  & KNN      & 1.67 \scriptsize(2.83)          & 3.22 \scriptsize(4.38)           & 5.92 \scriptsize(2.20)          & 5.35 \scriptsize(2.03)          & 5.23 \scriptsize(2.03)          \\
\cmidrule(lr){1-7}

\multirow{3}{*}{20} 
  & PedNStream (w/o OD) & 3.00 \scriptsize(2.16)          & 7.86 \scriptsize(5.08)           & \textbf{3.75 \scriptsize(1.33)} & \textbf{3.87 \scriptsize(1.33)} & \textbf{3.27 \scriptsize(1.18)} \\
  & PedNStream (w/ OD)  & \textbf{4.00 \scriptsize(3.87)} & \textbf{9.43 \scriptsize(6.40)}  & 3.89 \scriptsize(1.35)          & 4.02 \scriptsize(1.38)          & 3.44 \scriptsize(1.25)          \\
  & KNN      & 2.00 \scriptsize(2.08)          & 3.29 \scriptsize(3.04)           & 5.24 \scriptsize(0.92)          & 4.67 \scriptsize(0.81)          & 4.57 \scriptsize(0.82)          \\
\bottomrule
\end{tabular}
\end{table*}

As shown in \reftbl{tab:validation_results}, both variants of \emph{PedNStream} outperform KNN in volume ratio, NRMSE, and NDTW for every observation setting. The variant without OD calibration has the lowest NRMSE and NDTW throughout, indicating a better reconstruction of the temporal flow pattern than purely spatial interpolation. The OD-informed variant has higher GEH matching rates with 10 and 20 observed sensors, although it does not improve these rates with five sensors and slightly worsens NRMSE and NDTW. Overall, the results show that the network model can infer held-out aggregate counts from sparse observations more effectively than KNN..

\section{Closed-Loop Control and Runtime Analysis}\label{control_runtime}
\subsection{Crowd Management Case Study}
\begin{figure*}
    \centering
    \includegraphics[width=.8\textwidth]{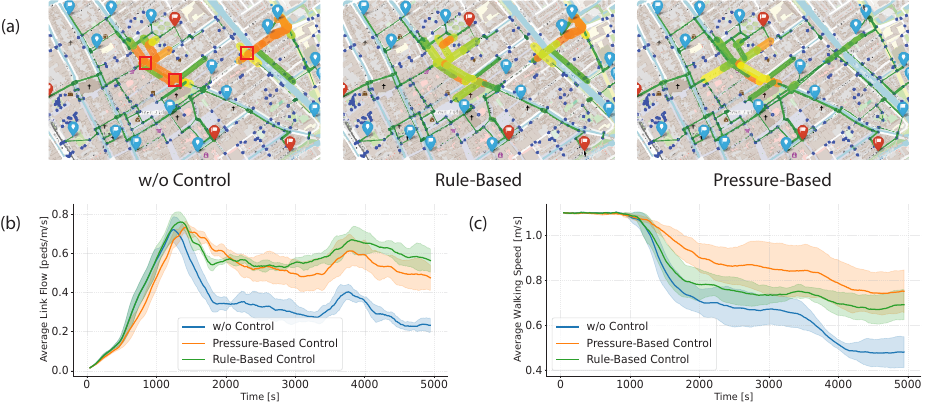}
    \caption{(a) Simulation snapshots with crowd management strategies at $t$=5000 s. Seven $Gater$ controllers are placed around the area highlighted by the red square. (b) The averaged link flow on the path \textit{0--8}. (c) The averaged walking speed on the path \textit{0--8}. The metrics are averaged over 5 random seeds.}
    \label{fig:control_delft}
\end{figure*}

To demonstrate the simulator's compatibility with closed-loop crowd management, we apply the $\textit{Gater}$ controllers introduced in \refsec{controller} to the Delft pedestrian network. The experiment contains 132 OD pairs and a festival-like demand pattern with two peaks around $t=1300$~s and $t=3700$~s. We place seven $Gaters$ around the area highlighted by the red square in \reffig{fig:control_delft}(a), where pedestrian interactions are most frequent. 

Each gater adjusts local gate widths based on observed link densities, thereby regulating inflow into the congested area and mitigating spillback at nearby intersections. This setup allows us to evaluate whether simple local feedback can improve network performance under strongly time-varying demand. 
%Specifically, we design two control algorithms one using basically rule-based logic to determine the gate width, another one utilize the concept of \emph{pressure}, e.g. the difference between the upstream and downstream density. The details of these two algorithms are introduced in Appendix \ref{apd}. 

We evaluate the control effect on path \textit{0--8} (yellow path at \reffig{delft_paths}), which traverses the regulated area and therefore reflects the local impact of the gating strategies. As shown in \reffig{fig:control_delft}(b) and \reffig{fig:control_delft}(c), both control strategies improve performance on this path. The rule-based controller yields the highest average link flow, whereas the pressure-based controller produces smoother behavior and a higher average walking speed. This difference is consistent with the algorithmic design, since the pressure-based method responds not only to local intersection density but also to upstream density conditions.

% the dashboard show the result of seed = 34

\subsection{Runtime and Scalability Analysis}
At each time step, \emph{PedNStream} first updates the state of every directed link. It then computes turning fractions and flow assignments for the nodes in the network's route sets. Let $E$ be the number of directed links, $P$ the number of active OD pairs, $\bar{N}$ the average number of nodes on the network's paths, $\bar{d}$ the average number of feasible incoming or outgoing links at a node, and $T$ the number of simulation time steps.

Updating the link states costs $\mathcal{O}(E)$ per time step. For each OD pair, the route choice and node flow calculations are performed along approximately $\bar{N}$ nodes. At a node, at most $\bar{d}^{2}$ movements between incoming and outgoing links are considered. If the number of candidate paths per OD pair is bounded and link attributes can be retrieved in constant time, the resulting cost per simulation run is $\mathcal{O}\!\left(T\left(E + P\bar{N}\bar{d}^{2}\right)\right)$.

This is an asymptotic estimate and omits constant implementation costs. Because $\bar{d}$ is usually small, and because the runtime experiment fixes the network topology and simulation horizon, the main changing term is the number of active OD pairs. Total pedestrian demand has little effect because demand is represented as aggregate flows rather than as individual agents. The dominant memory requirement is $\mathcal{O}(T \cdot E)$, as each link stores flow, speed, and density arrays over the full simulation horizon.

\begin{figure}
    \centering
    \includegraphics[width=\columnwidth]{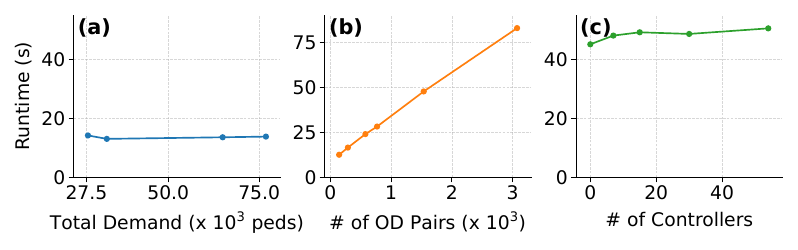}
    \caption{Runtime analysis of \emph{PedNStream}. (a): runtime with an increased demand profile, 20 OD pairs. (b): runtime with an increased number of OD pairs. (c): runtime with an increased number of controllers, with 132 OD pairs.}
    \label{runtime}
\end{figure}

We test these expectations by varying total demand, the number of OD pairs, and the number of controllers while keeping the network topology and simulation horizon fixed. The experiment is performed on a MacBook Pro with an Apple M1 Pro CPU, 16 GB RAM, and a 500 GB SSD. \reffig{runtime} shows that runtime changes little as total demand increases, but grows approximately linearly with the number of OD pairs. \reffig{runtime}(c) evaluates the additional cost of feedback gate control by varying the number of rule-based controllers. In the tested setting, controller updates add only modest overhead, indicating that \emph{PedNStream} can support feedback gate control experiments without a substantial loss of scalability.

% \section{Disscussion About the Usage of the Framework}
\section{Conclusion and Future Work} \label{conclusion}
% In this paper, we presented \emph{PedNStream}, a network-based pedestrian flow simulation tool built upon the original Link Transmission Model (LTM). The staged evaluation shows that the framework reproduces key pedestrian phenomena in synthetic networks. It also produces plausible large-scale dynamics consistent with observed counts and supports closed-loop crowd management at manageable runtime. Therefore, it offers a coherent bridge between descriptive pedestrian-network simulation and control-oriented experimentation.

% \emph{PedNStream} provides a foundation for developing real-time, large-scale crowd management algorithms and serves as a testbed for adapting traffic control methods originally designed for vehicular networks to pedestrian contexts. Future work will focus on integrating reinforcement learning and optimization-based controllers for automated crowd management, as well as validating the model using additional real-world datasets across diverse urban environments.

In this paper, we introduced \emph{PedNStream}, a network-based pedestrian simulation tool built upon the LTM. Staged evaluations show that the framework reproduces the designed flow mechanisms in synthetic networks. In a partially observed Melbourne reconstruction experiment, \emph{PedNStream} outperforms a spatial KNN baseline on several held-out aggregate-count metrics. The Delft case further demonstrates network-scale simulation, qualitative congestion patterns, controller integration, and manageable runtime. Together, these results position \emph{PedNStream} as a bridge between descriptive pedestrian-network simulation and control-oriented experimentation.

As a versatile testbed, \emph{PedNStream} provides a foundation for developing real-time, large-scale crowd management algorithms and for adapting macro-level traffic control methods to pedestrian contexts. Future work will focus on integrating reinforcement learning and optimization-based controllers for pedestrian traffic control algorithms, alongside validation against diverse real-world urban datasets with observed OD demand and route-choice information.

\appendices
\section{Software Architecture}\label{software_architecture}
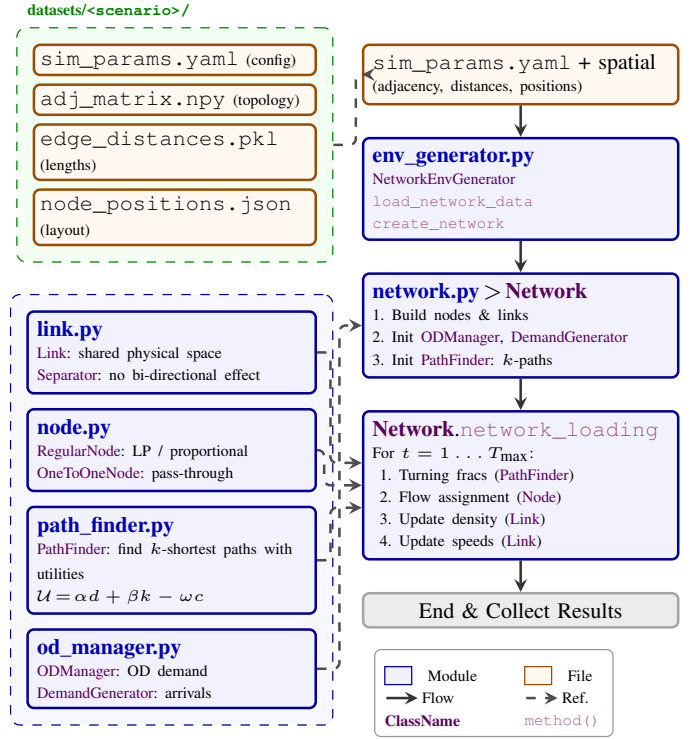
\begin{figure}[!t]
\centering
\resizebox{\columnwidth}{!}{%
\begin{tikzpicture}[
    node distance=0.35cm and 0.4cm,
    >=stealth,
    % --- Colors ---
    boxblue/.style={fill=blue!5, draw=blue!60!black},
    boxorange/.style={fill=orange!5, draw=orange!60!black},
    clscolor/.style={text=violet!70!black},
    methcolor/.style={text=magenta!60!black},
    modcolor/.style={text=blue!70!black},
    grcolor/.style={text=green!50!black},
    % --- Box Styles ---
    flowbox/.style={thick, rounded corners=2pt, inner sep=3pt, align=left,
                    boxblue, text width=3.3cm, font=\scriptsize},
    filebox/.style={thick, rounded corners=2pt, inner sep=3pt, align=left,
                    boxorange, text width=3.3cm, font=\scriptsize},
    sidebox/.style={thick, rounded corners=2pt, inner sep=3pt, align=left,
                    boxblue, text width=3.0cm, font=\scriptsize},
    sidefile/.style={thick, rounded corners=2pt, inner sep=2pt, align=left,
                     boxorange, text width=3.0cm, font=\scriptsize},
    endbox/.style={thick, rounded corners=2pt, inner sep=3pt, align=center,
                   fill=gray!15, draw=gray!70, text width=3.3cm, font=\scriptsize},
    databg/.style={draw=green!50!black, dashed, rounded corners=3pt,
                   fill=green!5, inner sep=5pt},
    modulebg/.style={draw=blue!50!black, dashed, rounded corners=3pt,
                     fill=blue!3, inner sep=5pt},
    % --- Arrows ---
    myarrow/.style={->, thick, draw=black!80},
    dasharrow/.style={->, thick, dashed, draw=black!70, rounded corners=2pt},
]

% ===== RIGHT: Main Flow (Anchors) =====

\node[filebox] (inputfiles) {
  \texttt{sim\_params.yaml} + spatial\\[-1pt]
  {\tiny (adjacency, distances, positions)}
};

\node[flowbox, below=0.35cm of inputfiles] (envloader) {
  \textbf{\textcolor{blue!70!black}{env\_generator.py}}\\[-1pt]
  {\tiny \textcolor{violet!70!black}{NetworkEnvGenerator}}\\[-1pt]
  {\tiny \textcolor{magenta!60!black}{\texttt{load\_network\_data}}}\\[-1pt]
  {\tiny \textcolor{magenta!60!black}{\texttt{create\_network}}}
};
\draw[myarrow] (inputfiles) -- (envloader);

\node[flowbox, below=0.35cm of envloader] (netinit) {
  \textbf{\textcolor{blue!70!black}{network.py}} $\!>\!$ \textbf{\textcolor{violet!70!black}{Network}}\\[-1pt]
  {\tiny 1. Build nodes \& links}\\[-1pt]
  {\tiny 2. Init \textcolor{violet!70!black}{ODManager}, \textcolor{violet!70!black}{DemandGenerator}}\\[-1pt]
  {\tiny 3. Init \textcolor{violet!70!black}{PathFinder}: $k$-paths}
};
\draw[myarrow] (envloader) -- (netinit);

\node[flowbox, below=0.35cm of netinit] (simrun) {
  \textbf{\textcolor{violet!70!black}{Network}}.\textcolor{magenta!60!black}{\texttt{network\_loading}}\\[-1pt]
  {\tiny For $t = 1 \ldots T_\text{max}$:}\\[-1pt]
  {\tiny ~1. Turning fracs (\textcolor{violet!70!black}{PathFinder})}\\[-1pt]
  {\tiny ~2. Flow assignment (\textcolor{violet!70!black}{Node})}\\[-1pt]
  {\tiny ~3. Update density (\textcolor{violet!70!black}{Link})}\\[-1pt]
  {\tiny ~4. Update speeds (\textcolor{violet!70!black}{Link})}
};
\draw[myarrow] (netinit) -- (simrun);

\node[endbox, below=0.35cm of simrun] (endnode) {End \& Collect Results};
\draw[myarrow] (simrun) -- (endnode);

% ===== LEFT TOP: Data Files =====

\node[sidefile, left=0.5cm of inputfiles.north west, anchor=north east] (f1) {
  \texttt{sim\_params.yaml} {\tiny (config)}
};
\node[sidefile, below=0.08cm of f1] (f2) {
  \texttt{adj\_matrix.npy} {\tiny (topology)}
};
\node[sidefile, below=0.08cm of f2] (f3) {
  \texttt{edge\_distances.pkl} {\tiny (lengths)}
};
\node[sidefile, below=0.08cm of f3] (f4) {
  \texttt{node\_positions.json} {\tiny (layout)}
};

\begin{scope}[on background layer]
  \node[databg, fit=(f1)(f2)(f3)(f4)] (databox) {};
  \node[anchor=south west, grcolor, font=\tiny\bfseries] at (databox.north west)
       {datasets/\texttt{<scenario>/}};
\end{scope}

\draw[dasharrow] (databox.east) -- ++(0.25,0) |- (inputfiles.west);

% ===== LEFT BOTTOM: LTM Modules =====

\node[sidebox, left=0.5cm of netinit.west, anchor=east, yshift=-0.3cm] (linkmod) {
  \textbf{\textcolor{blue!70!black}{link.py}}\\[-1pt]
  {\tiny \textcolor{violet!70!black}{Link}: shared physical space}\\[-1pt]
  {\tiny \textcolor{violet!70!black}{Separator}: no bi-directional effect}
};

\node[sidebox, below=0.15cm of linkmod] (nodemod) {
  \textbf{\textcolor{blue!70!black}{node.py}}\\[-1pt]
  {\tiny \textcolor{violet!70!black}{RegularNode}: LP / proportional}\\[-1pt]
  {\tiny \textcolor{violet!70!black}{OneToOneNode}: pass-through}
};

\node[sidebox, below=0.15cm of nodemod] (pathmod) {
  \textbf{\textcolor{blue!70!black}{path\_finder.py}}\\[-1pt]
  {\tiny \textcolor{violet!70!black}{PathFinder}: find $k$-shortest paths with utilities}\\[-1pt]
  {\tiny $\mathcal{U}\!=\!\alpha d + \beta k - \omega c$}
};

\node[sidebox, below=0.15cm of pathmod] (odmod) {
  \textbf{\textcolor{blue!70!black}{od\_manager.py}}\\[-1pt]
  {\tiny \textcolor{violet!70!black}{ODManager}: OD demand}\\[-1pt]
  {\tiny \textcolor{violet!70!black}{DemandGenerator}: arrivals}
};

\begin{scope}[on background layer]
  \node[modulebg, fit=(linkmod)(nodemod)(pathmod)(odmod)] (ltmbox) {};
\end{scope}

% Dashed arrows from modules to flow (flowing East to West now)
\draw[dasharrow] (linkmod.east) -- ++(0.15,0) |- ([yshift=0.25cm]simrun.west);
\draw[dasharrow] (nodemod.east) -- ++(0.08,0) |- (simrun.west);
\draw[dasharrow] (pathmod.east) -- ++(0.15,0) |- ([yshift=-0.25cm]simrun.west);
\draw[dasharrow] (odmod.east)   -- ++(0.25,0) |- (netinit.west);

% ===== LEGEND =====
% Anchored to align directly under the right column (endnode)
\node[below=0.4cm of endnode.south west, anchor=north west] (lstart) {};

% --- Row 1 ---
\node[right=0cm of lstart, fill=blue!5, draw=blue!60!black,
      minimum width=0.3cm, minimum height=0.2cm, inner sep=0pt] (lb1) {};
\node[right=0.05cm of lb1, font=\tiny] (lt1) {Module};

% Increased gap to 0.4cm to accommodate wider text in row 3
\node[right=0.4cm of lt1, fill=orange!5, draw=orange!60!black,
      minimum width=0.3cm, minimum height=0.2cm, inner sep=0pt] (lb2) {};
\node[right=0.05cm of lb2, font=\tiny] (lt2) {File};

% --- Row 2 ---
\node[below=0.25cm of lb1.west, anchor=west, inner sep=0pt] (la1s) {};
\draw[->, thick, draw=black!80] (la1s.center) -- ++(0.35,0)
      node[right, font=\tiny, inner sep=1pt] (lt5) {Flow};

\node[below=0.25cm of lb2.west, anchor=west, inner sep=0pt] (la2s) {};
\draw[->, thick, dashed, draw=black!70] (la2s.center) -- ++(0.35,0)
      node[right, font=\tiny, inner sep=1pt] (lt6) {Ref.};

% --- Row 3 ---
% Align Class Name directly under the Flow arrow
\node[below=0.25cm of la1s.west, anchor=west, inner sep=0pt, font=\tiny\bfseries, text=violet!70!black] (lt7) {ClassName};

% Align Method directly under the Ref arrow
\node[below=0.25cm of la2s.west, anchor=west, inner sep=0pt, font=\tiny, text=magenta!60!black] (lt8) {\texttt{method()}};

% --- Bounding Box ---
\begin{scope}[on background layer]
  \node[draw=gray!80, rounded corners=2pt, inner sep=3pt,
        fill=white, fit=(lb1)(lt1)(lb2)(lt2)(la1s)(lt5)(la2s)(lt6)(lt7)(lt8)] {};
\end{scope}

\end{tikzpicture}%
}
\vspace{2pt}
\caption{PedNStream architecture and simulation workflow. It consists of three main components: the data and configuration layer, core LTM modules for network construction, and the execution pipeline for network loading.}
\label{fig:pednstream_workflow}
\end{figure}

\reffig{fig:pednstream_workflow} summarizes the software modules and execution workflow of \emph{PedNStream}. Each scenario is stored under \path{datasets/<scenario_name>/}. The file \texttt{sim\_params.yaml} contains the model and controller parameters, while separate files define the adjacency matrix, link lengths, and node positions. During initialization, these inputs are converted into network, OD demand, and candidate path objects. Separating the scenario files from the model logic supports reproducible experiments and allows a network to be reconfigured without changing the simulation code.

The \texttt{Network} object coordinates the execution loop. At each time step, the routing module updates turning fractions, the node module assigns transfer flows, and the link module updates flow, density, and speed. Controllers access the network state through a common interface and modify only their assigned control variables. This organization allows routing models, node solvers, link models, and controllers to be replaced independently.

\section{Rule-Based and Pressure-Based Gate Controllers}\label{apd}
We implement two baseline controllers that operate on the same observation and
action interfaces. Both are decentralised: each junction
controller acts on its local set of gates $\mathcal{G}$ independently, using
only features observable at that junction. The two controllers differ in how
they map local congestion information to a gate-width adjustment.

\paragraph{Rule-Based Controller (Algorithm~\ref{alg:rule_based_gater}).}
The rule-based controller applies a discrete bang-bang update driven by a
density threshold $k^\star$. For every gate $g$, the own-direction density
$k(g)$ is compared against $k^\star$; when exceeded, the gate is closed
by a fixed step $\Delta w$. If neither direction exceeds its directional threshold but the total density of both directions exceeds the combined-density threshold, the controller closes the gate for the direction with the higher density. On the other hand, when no congestion signal is
detected, the gate is opened by $\Delta w$. This controller is simple and
interpretable, but its fixed step size and hard threshold make it
oscillation-prone near $k^\star$ and insensitive to the magnitude of
congestion.

\begin{algorithm}[H]
\caption{Rule-Based Gate Control}
\label{alg:rule_based_gater}
\begin{algorithmic}[1]
\Require Set of controlled gates $\mathcal{G}$ at junction $u$; density threshold $k^\star$; fixed step $\Delta w$
\Require For each $g \in \mathcal{G}$: own-direction density $k(g)$, paired density $k'(g)$, and current width $w(g)$
\Procedure{RuleBasedControl}{$\mathcal{G}, k^\star, \Delta w$}
    \For{each gate $g \in \mathcal{G}$}
        \If{$k(g) > k^\star$} 
        % \Comment{own direction congested $\rightarrow$ close}
            \State $a(g) \gets w(g) - \Delta w$
        \ElsIf{$k(g) + k'(g) > k^\star \;\land\; k(g) \geq k'(g)$} 
        % \Comment{link congested, own side dominates}
            \State $a(g) \gets w(g) - \Delta w$
        \Else \Comment{safe to open the gate}
            \State $a(g) \gets w(g) + \Delta w$
        \EndIf
    \EndFor
    \EndProcedure
\Ensure Target gate widths $a(g)$ for all $g \in \mathcal{G}$
\end{algorithmic}
\end{algorithm}

\paragraph{Pressure-Based Controller (Algorithm~\ref{alg:pressure_based_gater}).}
The pressure-based controller is inspired by max-pressure control from
vehicular traffic signal literature~\cite{varaiya2013max}. For each
gate $g$, it defines a local pressure
$P(g) = k_{\text{up}}(g) - k_{\text{down}}(g)$, i.e.\ the difference
between upstream demand and downstream resistance. A proportional law
$\delta(g) = \mathrm{clip}(K \cdot P(g),\, -\Delta_{\max}, +\Delta_{\max})$
translates pressure into a continuous, rate-limited width adjustment: the gate
opens aggressively under high positive pressure, closes under negative
pressure, and remains nearly stationary near equilibrium. Compared to the
rule-based variant, this controller is smoother, naturally scales its response
to the severity of imbalance, and requires no threshold tuning beyond the gain
$K$ and the safety clamp $\Delta_{\max}$.

Both controllers produce target gate widths that are subsequently
clipped to their physical bounds $[0, \text{width}(\ell)]$, ensuring feasibility regardless of the controller's raw
output. They therefore serve as lightweight, interpretable baselines against
which the benefits of learned, globally-coordinated policies can be measured.

\begin{algorithm}[H]
\caption{Pressure-Based Gate Control}
\label{alg:pressure_based_gater}
\begin{algorithmic}[1]
\Require Set of controlled gates $\mathcal{G}$ at junction $u$; proportional gain $K$; per-step bound $\Delta_{\max}$
\Require For each $g \in \mathcal{G}$: upstream density $k_{\text{up}}(g)$, downstream density $k_{\text{down}}(g)$, and current width $w(g)$
\Procedure{PressureControl}{$\mathcal{G}, K, \Delta_{\max}$}
    \For{each gate $g \in \mathcal{G}$}
        \State $P(g) \gets k_{\text{up}}(g) - k_{\text{down}}(g)$ 
        % \Comment{pressure = demand $-$ resistance}
        \State $\delta(g) \gets \mathrm{clip}\!\left(K \cdot P(g),\, -\Delta_{\max},\, +\Delta_{\max}\right)$
        \State $a(g) \gets w(g) + \delta(g)$
    \EndFor
    \EndProcedure
\Ensure Target gate widths $a(g)$ for all $g \in \mathcal{G}$
\end{algorithmic}
\end{algorithm}

\bibliographystyle{IEEEtran}
\bibliography{references}

\end{document}